\documentclass{article}

\usepackage{enumerate}%
\usepackage{microtype}
\usepackage{graphicx}
\usepackage{wrapfig}
\usepackage{subcaption}
\usepackage{standalone}
\usepackage{booktabs} %

\usepackage{siunitx} %
\usepackage{multirow}
\usepackage{makecell}
\usepackage{caption}
\usepackage{paralist}

\usepackage{amsmath}

\usepackage{hyperref}

\usepackage{csquotes}
\usepackage[capitalize, nameinlink]{cleveref}

\usepackage[accepted]{icml2022}

\usepackage{amsmath,amsfonts,bm}

\def\eqref#1{equation~\ref{#1}}

\def\1{\bm{1}}

\def\rvs{{\mathbf{s}}}

\def\rvx{{\mathbf{x}}}

\def\rvz{{\mathbf{z}}}

\def\rmS{{\mathbf{S}}}

\def\rmX{{\mathbf{X}}}

\def\rmZ{{\mathbf{Z}}}

\def\vg{{\bm{g}}}
\def\vh{{\bm{h}}}

\def\vw{{\bm{w}}}
\def\vx{{\bm{x}}}
\def\vy{{\bm{y}}}
\def\vz{{\bm{z}}}

\def\mS{{\bm{S}}}

\def\mW{{\bm{W}}}
\def\mX{{\bm{X}}}

\def\mZ{{\bm{Z}}}

\DeclareMathAlphabet{\mathsfit}{\encodingdefault}{\sfdefault}{m}{sl}
\SetMathAlphabet{\mathsfit}{bold}{\encodingdefault}{\sfdefault}{bx}{n}

\newcommand{\R}{\mathbb{R}}

\newcommand{\sigmoid}{\sigma}

\renewrobustcmd{\bfseries}{\fontseries{b}\selectfont}

\usepackage{xcolor}

\usepackage{xspace}

\newcommand{\needcite}[1][]{\textbf{\color{red} [cite\ifthenelse{\equal{#1}{}}{}{~{#1}}]}\xspace}

\newcommand{\msg}[1]{{\color{blue} [#1]}}
\newcommand{\edit}[2][]{{\textcolor{orange}{\sout{#2}}}\ifthenelse{\equal{#1}{}}{}{\ifthenelse{\equal{#2}{}}{}{$\rightarrow$}{\textcolor{green!40!black!90}{#1}}}}

\usepackage{amsfonts}
\usepackage{amsmath}
\usepackage{amsthm}
\usepackage{mathtools}
\usepackage{ulem}

\usepackage{ifthen}

\renewcommand{\epsilon}{\varepsilon}

\newcommand{\seq}[3]{\left\{#1_{#2}\right\}_{\ifx&#3&\else #2=1\fi}^{#3}}  %
\newcommand{\range}[2][]{{\ifx&#1&1, 2, \dots, #2\else#1_1, #1_2,\dots, #1_{#2}\fi}}

\newcommand{\norm}[2][]{{\left|\left|#2\right|\right|}_{#1}}

\newcommand{\spd}[2][]{\partial^{#1}_{#2}}

\DeclarePairedDelimiterX{\infdivx}[2]{(}{)}{%
	#1 \delimsize\|\, #2%
}

\DeclareMathOperator{\KLoperator}{KL}
\newcommand{\KLop}[2]{\KLoperator\infdivx{#1}{#2}}

\DeclareMathOperator{\Eoperator}{\mathbb{E}}
\newcommand{\Eop}[2][]{\Eoperator_{#1}\left[{#2}\right]}

\newcommand{\numberthis}{\addtocounter{equation}{1}\tag{\theequation}}

\theoremstyle{definition}

\usepackage{amsthm}



\usepackage{pifont}

\newcommand{\loss}{L}

\newcommand{\etab}{{\boldsymbol{\eta}}}

\newcommand{\latentsize}{l}

\usepackage[toc,page,header]{appendix}
\newcommand{\checkbox}{\makebox[0pt][l]{$\openbox$}\raisebox{.15ex}{\hspace{0.1em}$\checkmark$}} %

\colorlet{darkgreen}{green!50!black!}

\makeatletter
\newcommand{\subalign}[1]{%
  \vcenter{%
    \Let@ \restore@math@cr \default@tag
    \baselineskip\fontdimen10 \scriptfont\tw@
    \advance\baselineskip\fontdimen12 \scriptfont\tw@
    \lineskip\thr@@\fontdimen8 \scriptfont\thr@@
    \lineskiplimit\lineskip
    \ialign{\hfil$\m@th\scriptstyle##$&$\m@th\scriptstyle{}##$\hfil\crcr
      #1\crcr
    }%
  }%
}
\makeatother

\usepackage{url}

\usepackage{array}   %
\newcolumntype{L}{>{$}l<{$}} %
\newcolumntype{C}{>{$}c<{$}} %
\newcolumntype{R}{>{$}r<{$}} %

\usepackage{etoolbox}

\newif\ifcomments
\newif\ifdraft

\renewcommand{\msg}[3]{{\ifcomments\textcolor{blue}{\textbf{#1.} #2}\else\ifstrempty{#3}{\textcolor{blue}{\textbf{#1.} #2}}{\ifdraft\textcolor{blue}{\textbf{#1}}\newline#3\else#3\fi}\fi}}

\newcommand{\pages}[1]{\ifdraft\hfill\textcolor{blue}{\textbf{#1} pages}\fi}

\makeatletter
\DeclareRobustCommand\onedot{\futurelet\@let@token\@onedot}
\def\@onedot{\ifx\@let@token.\else.\null\fi\xspace}

\newcommand{\ie}{i.e\onedot}
\newcommand{\eg}{e.g\onedot}
\newcommand{\wrt}{w.r.t\onedot}
\newcommand{\randomvariable}{r.v\onedot}
\makeatother

\newcommand{\ours}{ours\xspace}

\newcommand{\problem}{modality collapse\xspace}
\newcommand{\Problem}{Modality collapse\xspace}
\newcommand{\block}{impartiality block\xspace}
\newcommand{\Block}{Impartiality block\xspace}

\crefformat{section}{\S#2#1#3}
\crefmultiformat{section}{\S\S#2#1#3}{ and~#2#1#3}{, #2#1#3}{, and~#2#1#3}
\creflabelformat{equation}{#2#1#3}  %
\crefformat{algorithm}{Alg.~#2#1#3}

\usepackage{xifthen}
\newcounter{mylabelcounter}

\makeatletter
\newcommand{\acdef}[2][]{%
\ifthenelse{\equal{#1}{}}{%
    \refstepcounter{mylabelcounter}#2%
        \immediate\write\@auxout{%
          \string\newlabel{ac:#2}{{1}{\thepage}{{\unexpanded{#2}}}{mylabelcounter.\number\value{mylabelcounter}}{}}%
        }%
    }{%
    \refstepcounter{mylabelcounter}#2 (#1)%
        \immediate\write\@auxout{%
          \string\newlabel{ac:#1}{{1}{\thepage}{{\unexpanded{#1}}}{mylabelcounter.\number\value{mylabelcounter}}{}}%
        }%
    }%
}
\makeatother

\usepackage{suffix}
\newcommand\ac[1]{\nameref{ac:#1}\xspace}
\WithSuffix\newcommand\ac*[1]{{\hypersetup{linkcolor=black}\nameref{ac:#1}\xspace}}

\newcounter{phase}
\newcommand{\nop}{\ignorespaces}
\newcommand{\orange}[1]{\textcolor{orange}{#1\xspace}}

\newcommand{\added}[2][1]{\ifthenelse{\value{phase}=#1}{\orange{#2}}{#2}}
\newcommand{\removed}[2][1]{\ifthenelse{\value{phase}=#1}{\sout{#2}}{\nop}}
\newcommand{\replaced}[3][1]{\ifthenelse{\value{phase}=#1}{\sout{#2} \orange{$\rightarrow$ #3}}{#3}}

\AtBeginDocument{\colorlet{defaultcolor}{.}}
\newenvironment{added*}[1][1]{\ifthenelse{\value{phase}=#1}{\color{orange}}{}}{\color{defaultcolor}}

\icmltitlerunning{Mitigating Modality Collapse in Multimodal VAEs}

\begin{document}

\twocolumn[

\icmltitle{Mitigating Modality Collapse in Multimodal VAEs \\ via Impartial Optimization}

\icmlsetsymbol{equal}{*}

\begin{icmlauthorlist}
\icmlauthor{Adri\'an Javaloy}{to}
\icmlauthor{Maryam Meghdadi}{to}
\icmlauthor{Isabel Valera}{to,goo}
\end{icmlauthorlist}

\icmlaffiliation{to}{Department of Computer Science, Saarland University, Germany}
\icmlaffiliation{goo}{MPI for Software Systems, Saarland, Germany}

\icmlcorrespondingauthor{Adri\'an Javaloy}{ajavaloy@cs.uni-saarland.de}

\icmlkeywords{variational autoencoders,modality collapse,multitask learning,conflicting gradients,negative transfer}

\vskip 0.3in
]

\printAffiliationsAndNotice{}  %

\begin{abstract}

A number of variational autoencoders (VAEs) have recently emerged with the aim of modeling multimodal data, \eg, to jointly model  images and their corresponding captions.
Still,  multimodal VAEs tend to focus solely on a subset of the modalities, \eg, by fitting the image while neglecting the caption. We refer to this limitation as \textit{\problem}.
In this work, we  argue that this effect is a consequence of \textit{conflicting gradients} during multimodal VAE training. 
We show how to detect the sub-graphs in the computational graphs where gradients conflict (\textit{\block{s}}), as well as  how to leverage existing gradient-conflict solutions from multitask learning to mitigate \problem. That is, to ensure \textit{impartial optimization} {across modalities}. 
We apply our training framework to several multimodal VAE models, losses and datasets from the literature, and empirically show  that our framework  significantly improves the reconstruction performance, conditional generation, and coherence of the latent space across modalities.
\end{abstract}

\section{Introduction\pages{1}}

Variational autoencoders (VAEs)~\citep{diederik2014auto} enjoy great success in domains such as images, text, and temporal data~\citep{nvidia-vae, xu2017variational-text, Mehrasa_2019_CVPR}.
Their application to multimodal data, \eg, to model images and their captions, remains a challenge, since models tend to accurately fit only a subset of the modalities, neglecting the rest.  We here refer to this problem as \textit{\problem}.

To overcome this issue, a number of tailored VAE models for tabular~\citep{nazabal2020hivae, NEURIPS2020vaem} and multimodal data~\citep{shi2019mmvae, sutter2021mopoe} have emerged over the years.
Interestingly, \citet{nazabal2020hivae} hypothesized that \problem is a result of to disparities between gradients across  modalities during training.

Following this inkling, we study \problem as a result of gradient conflicts in specific blocks of the computational graph, which we here call \block{s}~(\cref{sec:mm-and-optim}). 
To address this problem, we propose a multimodal VAE training pipeline, which leverages existing multitask learning solutions~(\cref{sec:dealing-with}) to favor an impartial optimization process that does not favor a subset of modalities over the rest.
We show the flexibility of our approach by applying our pipeline to several existing VAE models previously proposed in the literature to fit multimodal and tabular data~(\cref{sec:tailored-models}). 
Our empirical results on different datasets, models and training losses~(\cref{sec:experiments}) show that impartial optimization results in a more accurate fit of the marginal, joint  and conditional distributions over all modalities.

\msg{Notation}{Explain notation we use (such as $\mZ_{1:K}$).}{
\textbf{Notation.} 
We use the set indexing notation $\vx_A$, where $A$ is a set of indexes, \eg, $\rmZ_{1:K}$ denotes a sequence from $1$ to $K$.
We denote by $\mathbf{1}$ a vector full of ones, and by $[ \cdot ]$ the concatenation operator.
$\mathcal{P}(D)$ denotes the power-set of $D$ elements, and $|\mathcal{A}|$ the number of elements of the set $\mathcal{A}$.
}

\section{Preliminaries\pages{0.75}}

\textbf{Multimodal data.} In this work, we consider multimodal data, \ie, data coming from different sources and/or forms. %
Specifically, we consider as input data i.i.d. samples from a multimodal random variable (\randomvariable) \hbox{$\rmX = \begin{bmatrix} \range[\rvx]{D} \end{bmatrix}$}, where the $d$-th modality is fully described by the \randomvariable $\rvx_d$.
Note that we do not make any assumptions on the modalities, allowing for $\rvx_d$ of different sizes (\eg, images and their labels) and types (\eg, continuous vs. discrete).
We refer to $\rmX$ as heterogeneous when each modality $\rvx_d$ in $\rmX$ is unidimensional and can be of a different statistical type (\eg, normal or categorical).
Thus, we consider heterogeneous data as a special case of multimodal data.
Notice however that in the literature heterogeneous data is often studied independently (\eg, \citet{nazabal2020hivae, NEURIPS2020vaem}) of multimodal problems, and it is promiment in applications that deal with real-world tabular data.

\textbf{Variational autoencoders} (VAEs)~\citep{diederik2014auto} are probabilistic models that learn to model the data by assuming the existence of some latent variable $\rmZ$. Specifically, they learn the likelihood function that best approximates the input (decoder), $p_\theta(\rmX|\rmZ)$, and an approximation to the posterior distribution of $\mZ$ (encoder), $q_\phi(\rmZ|\rmX)$. %
During learning, VAEs maximize a function of the following form:
\begin{equation}
    \loss(\theta, \phi) = \Eop[\rmX]{\Eop[\rmZ_{1:K}\sim q_\phi]{\log\sum_{k=1}^K\frac{p_\theta(\rmX,\rmZ_k)}{q_\phi(\rmZ_k|\rmX)}}}, \label{eq:iwae}
\end{equation}  %
where $p_\theta(\rmX,\rmZ) = p_\theta(\rmX|\rmZ)p(\rmZ)$, and $\rmZ_{1:K}$ is an i.i.d sequence of length $K$. 
This formulation includes the original ELBO~\citep{diederik2014auto}, as well as the importance weighted loss (IWAE) from~\citet{burda2016importance}.

One important detail here is that the functional form of $p_\theta$ (and $q_\phi$) is usually fixed beforehand---\eg, as a normal distribution---while a neural network determines its parameters $\etab$. 
Importantly, when dealing with multimodal data, the usual practice is to assume that the likelihood fully factorizes accross modalities, \ie, 
\begin{equation}
    p_\theta(\rmX|\rmZ) = \prod_{d=1}^D p_d(\rvx_d;~\etab_d(\rmZ; \theta)), \label{eq:factorization}
\end{equation}
where $p_d$ accounts for the statistical properties of  $\rvx_d$.

\subsection{State-of-the-art\pages{0.5}}

\textbf{Heterogeneous data:} 
The most prominent \added{VAE} models \added{found in the literature} are probably HI-VAE~\citep{nazabal2020hivae} (\added{see, }\cref{sec:hivae}), \added{originally} designed for missing data imputation \added{tasks}, and VAEM~\citep{NEURIPS2020vaem}\added{, designed instead} for active data acquisition \added{tasks}.
More recently, SHIVAE~\citep{shivae} has been \replaced{proposed to extend}{introduced as an extension of} HI-VAE \replaced{for}{to deal with} temporal data.

\textbf{Multimodal data:} 
\added{We focus in this work on} mixture-based VAE models (\added{see} \cref{sec:mixture-based})\added{, which} are at the moment an active area of research. While MVAE~\cite{Wu2018MVAE}, MMVAE~\cite{shi2019mmvae}, and MoPoE~\cite{sutter2021mopoe} are the models to beat, different extensions \added{compatible with our proposed framework} keep coming up, \eg, using alternative training functions~\cite{shi2021relating,sutter2020mmjsd}.
\added{For a survey on other multimodal methods refer to the work of, \eg, \citet{guo2019deep,baltruvsaitis2018multimodal}.}

\begin{figure*}[t]
	\centering
    \hfill %
    \begin{subfigure}[c]{.49\textwidth}
        \centering
        \includestandalone[width=\linewidth, mode=image|tex]{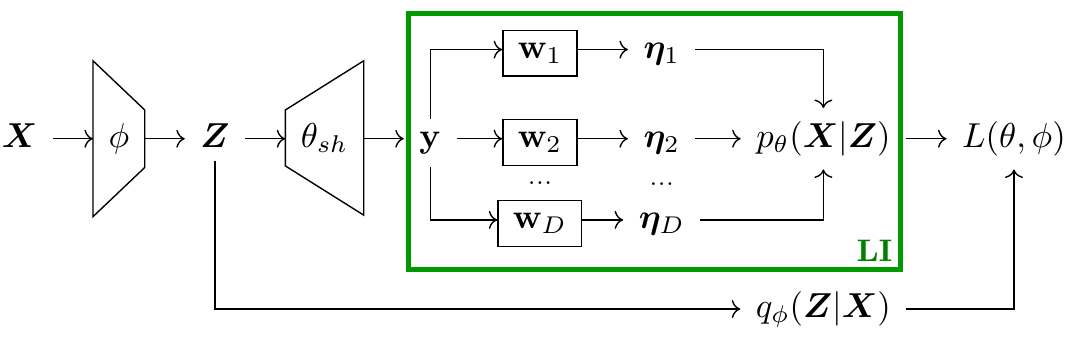}
        \caption{Forward pass.} \label{fig:vae-forward}
    \end{subfigure} %
    \hfill %
    \begin{subfigure}[c]{.49\textwidth}
        \centering
        \includestandalone[width=\linewidth, mode=image|tex]{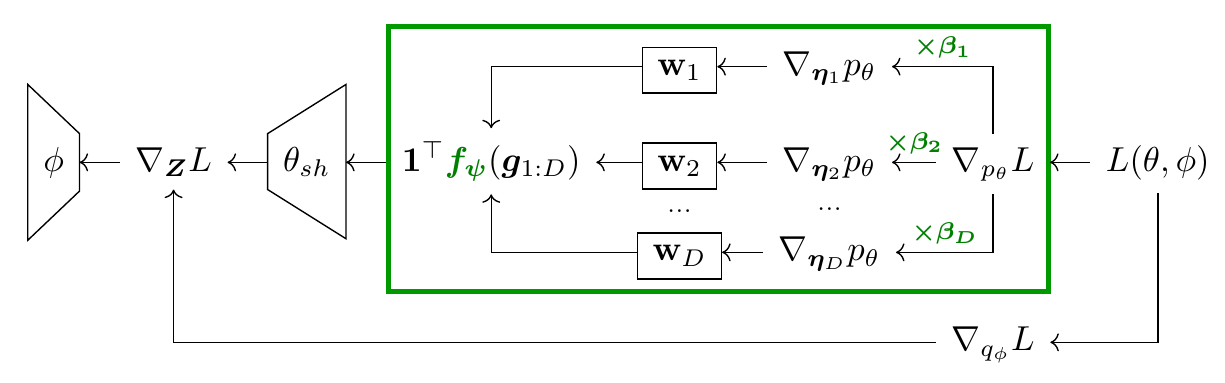}
        \caption{Modified backward pass.} \label{fig:vae-backward}
    \end{subfigure} %
    \caption{Schematic computational graph of a basic multimodal VAE: (a) forward pass, taking $\rmX$ as input and producing the training objective; (b) backward pass, modified to alleviate \problem (see \cref{sec:dealing-with}). The \block, which encloses conflicting gradients, as well as the modifications proposed in this work to tackle them, are highlighted in {green}.}
\end{figure*}

\section{Impartial Optimization in Multimodal VAEs\pages{2.5}} \label{sec:mm-and-optim}

In this section, we investigate the standard assumptions and goals of multimodal VAEs, as well as discuss the optimization challenges that cause \problem. Then, we propose a flexible learning approach to palliate this issue.

First, let us bring multimodal modeling to context. 
When we think of multimodal applications (\eg, missing data imputation, or joint data generation) these are tasks that involve not only explaining the different modalities in the data, but jointly capturing the interactions and dependencies between each pair of modalities. %
That is, the main goal  (often implicit) of  multimodal learning is thus to  accurately approximate the marginal, joint and conditional distributions over all modalities.

\textbf{\acdef[LI]{Likelihood Impartiality}:}  In order to do so, it is essential to accurately fit the likelihood of all modalities without neglecting any of them. \textit{We thus aim for a learning process that does not prioritize the learning of,  or equivalently, that  is impartial to, the likelihood of the different modalities. }

We argue here that the reason why likelihood impartiality is often not satisfied by multimodal VAE training is the computational graph resulting from the likelihood factorization in \cref{eq:factorization}. 
We illustrate this idea in \cref{fig:vae-forward}, where we highlight the problematic sub-graph in the computational graph, referred to as \textbf{\block}. 

As an example, assume here that the last layer of the decoder is a linear layer with parameter~$\mW$, and let us denote by $\theta_{sh}$ the rest of the decoder parameters, which are \textit{shared} across all modalities.
Then, we can write the likelihood parameters as $\etab = \sigmoid(\vy\mW)$, where $\vy$ is the output of the decoder up to the last shared layer, and $\sigmoid$ is an elementwise transformation to ensure that each parameter satisfies its distributional constraints (\eg, positive variance).
Making now the modality dependency explicit we can write, %
$\begin{bmatrix}\range[\etab]{D}\end{bmatrix} = \sigmoid(\vy\begin{bmatrix}\range[\vw]{D}\end{bmatrix})$, where it is now clear that all modalities share $\vy$, while the parameters $\vw_d$ are exclusive of the likelihood for the $d$-th modality.

An \block (green square in \cref{fig:vae-forward}) encloses a sub-graph in which a split-and-merge pattern across modalities appears, which we will recurrently observe later in \cref{sec:tailored-models}. %
In the forward pass, the \block takes a shared $\vy$ as {input}, which is \textit{independently} fed to each modality-specific ``head'' to compute %
$\etab_d$.
Then, these computations are collected to compute a common {output}, the total likelihood $p_\theta(\rmX|\rmZ)$.
Note that, outside this block, all computations are shared across modalities.

\Block{s} play an essential role on explaining \problem in multimodal VAEs.
First, we need to understand the effect of the split-and-merge pattern on the update rule of the shared parameters during optimization.
That is, we need to compute the gradient of $\loss(\theta, \phi)$ \wrt $\theta_{sh}$ (similar computations follow in the case of $\phi$), \textit{passing through the computational block}:%

\begin{align*}
    \nabla_{\theta_{sh}} \loss(\theta, \phi) &= \nabla_{\theta_{sh}} \vy\,\nabla_\vy \etab\,\nabla_\etab p_\theta\,\nabla_{p_\theta}\loss \\
	& = \nabla_{\theta_{sh}} \vy\textcolor{green!50!black}{\bm\left( \textcolor{black}{\sum_{d=1}^D \nabla_\vy \etab_d\,\nabla_{\etab_d} p_\theta}\bm\right)}\nabla_{p_\theta}\loss \\
	& = \nabla_{\theta_{sh}}\vy\, \sum_d \vg_d, \numberthis \label{eq:gradient}
\end{align*}
where $\vg_d\coloneqq\nabla_\vy \etab_d\,\nabla_{\etab_d} p_\theta\,\nabla_{p_\theta}\loss$ is the gradient of the loss \wrt $\vy$ through the $d$-th modality, as it is computed during back-propagation~\citep{Rumelhart1986backpropagation}.

\Cref{eq:gradient} reveals why \problem may occur during training.
Intuitively, each gradient $\vg_d$ represents the update direction that the model should follow to better explain the $d$-th modality. 
However, if there exist large discrepancies between different gradients $\vg_d$, \ie, in the presence of \textbf{conflicting gradients}, the overall gradient computation (namely, the sum $\sum_d \vg_d$) can benefit some modalities over others, leading to an update of the shared parameters that prioritize a subset of the modalities.

Therefore, our goal is to ensure impartiality across modalities in the computations that output the \block, such that no modality is neglected. Hence its name.
We remark here that the conflicting gradient problem are not exclusive to multimodal VAEs, and it has been studied in areas such as multitask learning (MTL).
Refer to \cref{sec:mtl-methods} for an overview of MTL. 
\subsection{Our Approach} 
\label{sec:dealing-with} %

In this section, we propose to modify the backward pass of the \block during training (since all outer computations are shared across modalities). We do so by  leveraging existing MTL solutions %
to enforce impartial optimization, and thus mitigate \problem.

We illustrate the proposed approach in \cref{fig:vae-backward} and \cref{alg:backward}, highlighting in green those parts that differ from usual back-propagation (see \cref{app:tailored-models} for a general formulation).
We propose two modifications within the \block to bring impartiality with respect to the modalities:
\begin{itemize}
\vspace{-15pt}
    \item  \textbf{Local step:} 
    Backpropagating through the heads, we re-weigh the gradients with respect to the likelihood parameters $\etab_d$ (which are local to each modality) by a factor of $\beta_d \in \R^+$ to keep them at a comparable scale. 
    We choose $\beta_d$ to be the number of dimensions of $\rvx_d$, similar to solutions in the literature (\eg, \citet{shi2019mmvae}).
    Note, however, that in prior work re-weighing was an ad-hoc fix in the forward pass (rather than in the backward pass), despite breaking probabilistic assumptions.\footnote{Specifically, that the likelihood integrates to one.}
    This step is also similar to loss balance in MTL. %
    Here,  we opt for a simple approach as it works well in practice, but more complex approaches could be also adapted to our framework, \eg, those proposed by~\citet{kendall2018multi}, \citet{Chennupati2019MultiNetMF}, and \citet{liu2021imtl}.
    \item \textbf{Global step.}
    Instead of propagating to the shared parameters ($\theta_{sh}$ and $\phi$) the gradient with respect to (the shared or global representation) $\vy$, we leverage existing MTL solutions to avoid conflicting gradients. %
    These solutions can be described as a (parameterized) function $f_\psi$ that takes a sequence of gradients $\vg_{1:D}$, and returns another of equal length $\tilde{\vg}_{1:D}\coloneqq f_\psi(\vg_{1:D})$, where the function $f_\psi$ is selected to mitigate conflicts (e.g., in magnitude or direction) in $\vg_{1:D}$. %
    We thus apply $f_\psi$ to the gradients with respect to $\vy$, and backpropagate $\sum_d \tilde{\vg}_d$ instead of $\sum_d \vg_d$. 
    Note that the function $f_\psi$ is determined by the specific MTL method that is applied. 
    \removed{For an overview of the considered MTL methods and their respective $f_\psi$, refer to \cref{sec:mtl-methods}.}
\end{itemize}
To sum up, we address  \problem within each \block by: i)~scaling local gradients \wrt $\etab_d$ by $\beta_d$ to make them comparable; and ii)~leveraging existing MTL solutions to modify the gradients \wrt $\vy$ such that they do not conflict, propagating this impartial gradients to the shared parameters.

There are two important remarks to make here. 
First, the local character of \block{s} is in stark contrast with traditional MTL: we do not make any assumption on the outer computational graph, nor the number of blocks in the graph.
Second, the optimal choice of $f_\psi$ depends on the problem setting, with no clear winner among existing MTL solutions.
Therefore, we treat the choice of algorithm $f_\psi$ as a hyperparameter, which we need to cross-validate.

\begin{algorithm}[tb]
	\caption{Backward pass within the \block.}
	\label{alg:backward}
	\begin{algorithmic}[1]
		\STATE {\bfseries Input:} Output gradient, $\nabla_{p_\theta} \loss$.
		\FOR{$d=1$ {\bfseries to} $D$}
		\STATE $\vh_d \gets \textcolor{green!50!black}{\bm {\beta_d}}  \nabla_{\eta_d} p_\theta\, \nabla_{p_\theta} \loss$
		\STATE $\nabla_{\omega_d} \loss \gets \nabla_{\omega_d} \etab_d\, \cdot \vh_d$
		\STATE $\vg_d \gets \nabla_\vy \etab_d\, \cdot \vh_d$
		\ENDFOR
		\STATE $\tilde{\vg}_{1:D} \gets \textcolor{green!50!black}{\bm {f_\psi}}(\vg_{1:D})$
		\STATE {\bfseries return} $\sum_d \tilde{\vg}_{d}$
	\end{algorithmic}
\end{algorithm}

\subsection{Conflicting Gradients Solutions}

\begin{added*}
Here, we briefly discuss the MTL solutions for conflicting gradients considered in the global step from \cref{sec:dealing-with}. Refer to \cref{sec:mtl-methods} for a full description. %

As explained above, these solutions modify the gradients through a function $f_\psi: \R^{D\times\latentsize} \rightarrow \R^{D\times\latentsize}$, where $\latentsize$ is the dimension of the latent variable $\rmZ$, and the input are the gradients for each task, stacked on the first dimension.

Moreover, they can be classified into two main categories, depending on the way they deal with conflicting gradients:
\begin{itemize}
    \item Scale-aware algorithms use a function $f_\psi$ that scales each gradient $\vg_d$ according to a given criterion, thus changing the magnitude of the gradient. That is, $f_\psi$ replaces each $\vg_d$ by $\omega_d\,\vg_d$.
    This type of solutions usually deal therefore with disparities in magnitude.
    
    \item Direction-aware algorithms, instead, attempt to fully homogenize task gradients. As a consequence, $f_\phi$ deals also with issues related with gradients pointing towards different directions of the parameter space, thus cancelling out each other when added up.
\end{itemize}

Note that the contribution of our work is to identify \textit{where} to modify gradients, rather than \textit{how} to modify them. Thus, as mentioned at the end of \cref{sec:dealing-with}, we cross-validate the choice of $f_\psi$ between different magnitude-aware~\citep{chen2017gradnorm,mgdaub,cagrad,liu2021imtl} and direction-aware~\citep{graddrop,yu2020pcgrad} options.
Moreover, our work is orthogonal to the choice of $f_\psi$, and therefore new algorithms can be easily included.
\end{added*}

\section{Extending Our Framework} 
\label{sec:tailored-models}  %

Next, we revisit different  VAE models proposed in the literature to handle multimodal data, and show how to apply them the ideas in \cref{sec:mm-and-optim} to avoid \problem.

\subsection{Heterogeneous VAE Models}
\label{sec:hivae}

\begin{figure}[t]
    \centering
    \includestandalone[width=\linewidth, mode=image|tex]{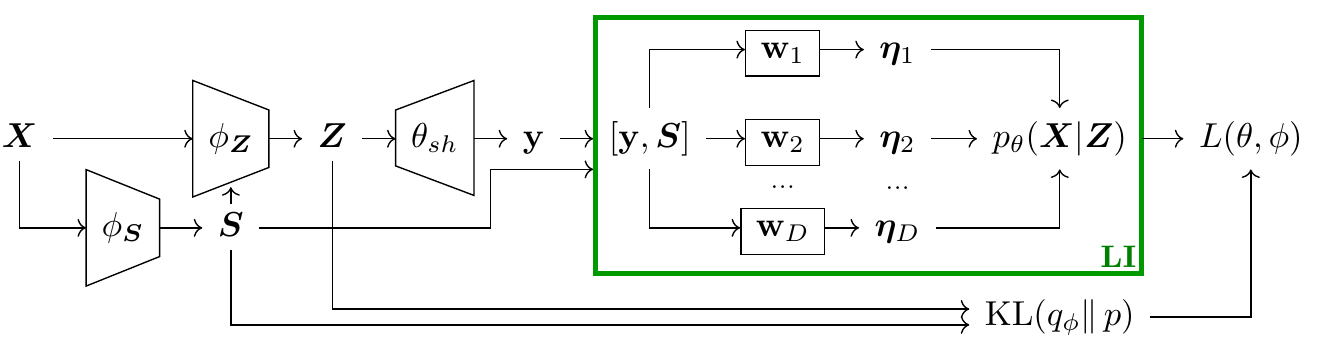}
    \caption{Forward pass of a \ac*{HI-VAE} and its \block.}
    \label{fig:forward-hivae}
\end{figure}

The \acdef[HI-VAE]{Heterogeneous-Incomplete VAE}~\citep{nazabal2020hivae} is a model specialized on handling heterogeneous data. %
While it differs from a standard VAE in several aspects---\eg, including a data normalization layer, its hierarchical structure  in the form of a Gaussian mixture prior is of especial interest to us. 
Quoting the original authors, this more expressive prior helps ``overcoming the limitations of having assumed a generative model that fully factorizes for every dimension'' (see \cref{eq:factorization}).

We show the computational graph of the HI-VAE in \cref{fig:forward-hivae}.
In short, HI-VAE introduces an additional latent variable, $\mS$, and defines the encoder (and prior) to be of the form $q_\phi(\mZ,\mS|\mX) = q_\phi(\mZ|\mS,\mX) q_\phi(\mS|\mX)$.
Akin to the example in~\cref{sec:mm-and-optim}, the last layer of the model is a linear layer, $\mW$, and the parameters are obtained as $\etab = \sigma([\vy, \mS]\mW)$. %
Note that \cref{eq:gradient} remains valid in this case. Moreover, there are additional conflicting-gradient problems, this time \wrt $\mS$:
\begin{align*}
    \nabla_{\phi_\rmS} p_\theta \nabla_{p_\theta}& \loss(\theta, \phi) = \nabla_{\phi_\rmS} \rmS\textcolor{green!50!black}{\bm\left( \textcolor{black}{\sum_{d=1}^D \nabla_\rmS \etab_d\,\nabla_{\etab_d} p_\theta}\bm\right)}\nabla_{p_\theta}\loss. \numberthis \label{eq:gradient-s}
\end{align*}

Eqs.~\ref{eq:gradient} and~\ref{eq:gradient-s} %
show that HI-VAE contains an \block with two different inputs, $\vy$ and $\rmS$.
Hence, we propose to tackle \problem by applying our approach (\cref{sec:dealing-with}), and thus \cref{alg:backward},  to both 
inputs. This implies using MTL twice, i.e., to learn $f_{\psi_\vy}$ and $f_{\psi_\rmS}$.

\subsection{Multimodal VAE Models} \label{sec:mixture-based}

\subsubsection{Mixture-based VAEs for Multimodal Data}

\begin{figure}[t]
    \centering
    \includestandalone[width=\linewidth, mode=image|tex]{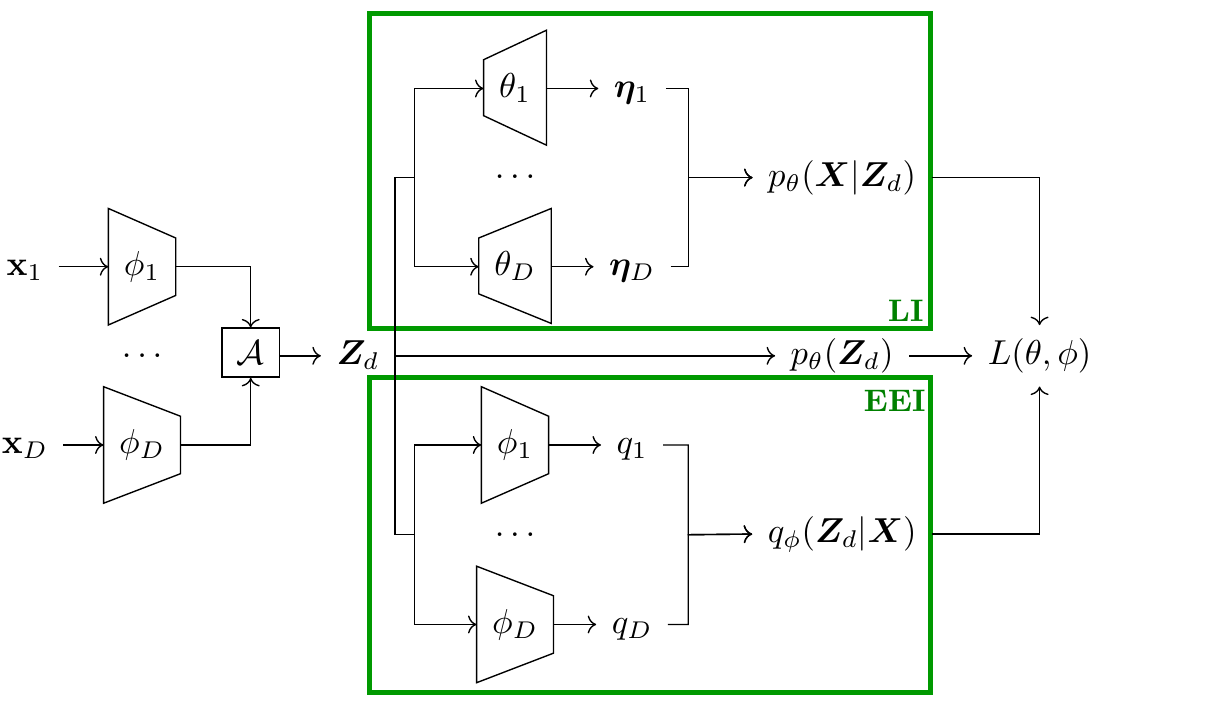}
    \caption{Forward pass of MMVAE, check \cref{fig:mmvae-theta} to see the blocks related with \ac{DEI}. Note that MMVAE only has unimodal experts, and that we only show a single latent sample $\rmZ_d$.}
    \label{fig:mvae-forward}
\end{figure}

One desirable property for multimodal VAEs is \textit{conditional generation}, \ie, sampling a modality having observed a different one, representing the same underlying concept. For example, sample the caption for a given image, or vice-versa. 
However, when the encoder is shared across all modalities, accurate conditional generation is not straight-forward.
Mixture-based multimodal VAEs solve this issue by introducing $D$ modality-exclusive encoders (and decoders), using
 as variational distribution a mixture model of the form:
\begin{equation}
    q_\phi(\rmZ|\rmX) = \frac{1}{|\mathcal{A}|} \sum_{A\in \mathcal{A}} q_A(\rmZ|\rmX_A), \label{eq:mixture}
\end{equation}
where $\mathcal{A}\subset\mathcal{P}(D)$ is a subset of all the possible combinations of modalities, and $q_A(\rmZ | \rmX_A)$ %
is an \textit{expert} composed of the modalities in $A\subset\{\range{D}\}$,  %
\begin{equation}
    q_A(\rmZ|\rmX_A) \propto \prod_{d\in A} q_{\phi_d} (\rmZ|\rvx_d).
\end{equation}

We can recover existing models by selecting different values for $\mathcal{A}$ (\Cref{fig:mvae-forward} illustrate  the forward pass of the MMVAE):

\begin{tabular}{ll}
    {MVAE~\citep{Wu2018MVAE}:} & $\mathcal{A} = \{\{\range{D}\}\}$, \\
    {MMVAE~\citep{shi2019mmvae}:} & $\mathcal{A} = \{ \{1\}, \dots, \{D\}  \}$, \\
    {MoPoE~\citep{sutter2021mopoe}:} & $\mathcal{A} = \mathcal{P}(D)$.
\end{tabular}

One setback of considering $q_\phi$ a mixture model is that we cannot longer differentiably sample from it.
First introduced by~\citet{shi2019mmvae}, and rediscovered by~\citet{Morningstar2021selbo}, we can overcome this issue by employing stratified sampling, leading to the following objective:
\begin{equation}
    \loss(\theta, \phi) = \sum_{A\in\mathcal{A}} \Eop[{\rmX,\rmZ_{1:K}^A}]{\log\sum_{k=1}^K\frac{p_\theta(\rmX,\rmZ_k^A)}{q_\phi(\rmZ_k^A|\rmX)}}.
    \label{eq:loose-loss}
\end{equation}
We refer to \cref{eq:loose-loss} as loose
since a tighter objective, \acdef{SIWAE}, can be derived%
~\citep{shi2019mmvae, Morningstar2021selbo}:
\begin{equation}
    \tilde{\loss}(\theta, \phi) = \Eop[\rmX,\left\{\rmZ_{1:K}^A\right\}_{\mathcal{A}}]{\log\sum_{A\in\mathcal{A}} \sum_{k=1}^K\frac{p_\theta(\rmX,\rmZ_k^A)}{q_\phi(\rmZ_k^A|\rmX)}}.
    \label{eq:tight-loss}
\end{equation}
Despite being tighter, this objective is notoriously known for suffering from \problem.
\citet{shi2019mmvae} discarded its use, showing empirical evidence of \problem and arguing that ``it leads to situations where the joint variational posterior collapses to one of the experts in the mixture.''

\subsubsection{Impartial Optimization} \label{sec:mixture-based-conflict}

Recall that our main goal is to accurately approximate the marginal, joint and conditional distributions over all modalities.
To achieve this objective, we now identify different \block{s} that may stray us from our goal.

Looking at \cref{fig:mvae-forward}, we find an upper \block, which corresponds once again of evaluating the factorized likelihood~(\cref{eq:factorization}).
For each expert $A$, we find such a \block, having each decoder as a head and its latent variable $\rmZ_A$ as the common input.
Hence, we can improve \ac{LI} by applying \cref{alg:backward} to each of these blocks.
Next, we focus on the specific problems of mixture-based models that may also contribute to \problem.
Just as in \cref{sec:mm-and-optim}, we first describe the goals to pursue in order to achieve conditional generation. Then, we study the parts of the computational graph that may hinder achieving these goals.

\textbf{\acdef[EEI]{Encoder Expert-Impartiality}:} 
In order to enable conditional generation, we need interchangeable encoders, so that we can replace them when modalities are missing. 
\textit{In other words, we need the ability to generate encoder samples that are impartial to the expert.}

Given the latent samples from an expert, $\rmZ_A$, we can compute how likely these samples are of coming from any another expert~$A'$ by computing $q_{A'}(\rmZ_A | \rmX_{A'})$.
Similar to the way $\vy$ could receive gradients from $p_\theta(\rmX|\rmZ)$ benefiting a subset of modalities (see \cref{sec:mm-and-optim}), $\rmZ_A$ can receive gradients from the mixture $q_\phi(\rmZ|\rmX)$ that favor a subset of modalities.
This \block can be observed in the bottom part of \cref{fig:mvae-forward}, as well as by computing the gradients of $L(\theta, \phi)$ \wrt $\rmZ_A$, passing through $q_\phi(\rmZ_A|\rmX)$, \ie:
\begin{align*}
    \nabla_{\phi_d} \rmZ_A &\nabla_{\rmZ_A} q_\phi \nabla_{q_\phi} \loss(\theta, \phi) = \\ 
    &= \nabla_{\phi_d} \rmZ_A \textcolor{green!50!black}{\bm\left( \textcolor{black}{\sum_{A'\in\mathcal{A}} \nabla_{\rmZ_A} q_{A'}}\bm\right)}\nabla_{q_\phi}\loss. \numberthis \label{eq:gradient-q}
\end{align*}

\Problem can thus appear as a consequence of conflicting gradients in \cref{eq:gradient-q}, having experts whose samples can only substitute a subset of other experts.
We can prevent it by applying \cref{alg:backward} to these \block{s}.

\textbf{\acdef[DEI]{Decoder Expert-Impartiality}:} Similar to \ac{EEI}, to have proper conditional generation we need interchangeable decoders that can generate their modality using any latent sample. 
\textit{That is, we aim for decoders that are impartial to the expert that generated the latent samples.}

\ac{DEI} relates to the passive role of the latent samples, where the decoder parameters\footnote{We do not consider the encoder parameters here, since we use the STL estimator~\citep{Roeder2017StickingTL}.} are optimized taking these samples as input.
In particular, each decoder $p_{\theta_d}(\rvx_d|\rmZ)$ is optimized to explain the \randomvariable $\rvx_d$ \textit{given the samples from each expert, $\rmZ_A \sim q_A(\rmZ|\rmX_A)$}, which is explicitly shown via stratification in \cref{eq:loose-loss,eq:tight-loss}.

This time, \problem would lead to decoders that can only generate their modality based on a subset of experts.
Building on the ideas from \cref{sec:mm-and-optim}, we can find that, for each decoder $p_{\theta_d}(\rvx_d|\rmZ)$, there exists an \block:
\begin{align*}
    \nabla_{\theta_d} \loss(\theta, \phi) &= \textcolor{green!50!black}{\bm\left( \textcolor{black}{\sum_{A\in\mathcal{A}} \nabla_{\theta_d} p_{\theta_d}^A\nabla_{p_{\theta_d}^A}\loss }\bm\right)}, \numberthis \label{eq:gradient-stratify}
\end{align*}
where we denote $p_{\theta_d}^A\coloneqq p_{\theta_d}(\rvx_d|\rmZ_A)$ for the sake of brevity.

Note that the \block in \cref{eq:gradient-stratify} (illustrated in \cref{fig:mmvae-theta} of \cref{app:tailored-models}) has as input $\theta_d$, the decoder parameters, and each sample $\rmZ_A$ as modality-specific head.
However, due to the flexibility offered by the \block{s}, we can reason and tackle \problem just as we did in the other cases: applying to each \block \cref{alg:backward}.

In total, there are  $2|\mathcal{A}| + D$ \block{s}  in a mixture-based VAE, for which we can use \cref{alg:backward} to palliate \problem. 
Extra details on their application can be found in \cref{app:tailored-models}.

\section{Experiments\pages{1.5 + 1.5}}\label{sec:experiments}

In this section, we assess the approaches shown in \cref{sec:mm-and-optim} and \cref{sec:tailored-models} for heterogeneous and multimodal settings.
All results shown here are averaged over 5 different seeds and bold numbers represent statistically significant values according to a one-sided Student's t-test ($\alpha = 0.1$), unless stated otherwise.
Additional details and  results can be found in \cref{app:models,app:experimental-setup}.

{
\sisetup{table-format=1.2, round-mode = places, round-precision = 2, detect-all}
\begin{table*}[t]
    \setlength\tabcolsep{2.4pt}
    \centering
    \caption{Test reconstruction errors (median over five seeds) for different datasets and VAE models. Statistically different values according to a corrected paired t-test ($\alpha=0.1$) are shown in bold. Models trained with our approach outperforms the baseline in most cases.} \label{tab:test_errors}
    {
        \begin{tabular}{cll|ccccccccc|ccc} 
            \toprule
            & & & \multicolumn{9}{c|}{Heterogeneous} & \multicolumn{3}{c}{Homogeneous} \\
            & & & {\textit{Adult}} & {\textit{Credit}} & {\textit{Wine}} & {\textit{Diam.}} & {\textit{Bank}} & {\textit{IMDB}} & {\textit{HI}} & {\textit{rwm5yr}} & {\textit{labour}}& {\textit{El Nino}}& {\textit{Magic}}& {\textit{BooNE}} \\ \midrule
            \multirow{7}{*}{\rotatebox[origin=c]{90}{{Standard VAE}}} & \multirow{2}{*}{{ELBO}} & vanilla & 0.213 & 0.128 & 0.086 & 0.187 & 0.203 & 0.082 & 0.170 & 0.105 & 0.109 & 0.109 & 0.064 & 0.042\\
            & & \ours & \textbf{0.104} & \textbf{0.041} & \textbf{0.071} & \textbf{0.139} & \textbf{0.043} & \textbf{0.032} & \textbf{0.041} & \textbf{0.026} & \textbf{0.063} & \textbf{0.068} & 0.058 & 0.039 \\ \cmidrule{2-15}
            & \multirow{2}{*}{{IWAE}} & vanilla & 0.226 & 0.134 & 0.075 & 0.185 & 0.199 & 0.090 & 0.155 & 0.094 & 0.098 & 0.086 & 0.053 & 0.037\\
            & & \ours & \textbf{0.129} & \textbf{0.051} & \textbf{0.066} & \textbf{0.125} & 0.076 & \textbf{0.035} & \textbf{0.042} & \textbf{0.032} & \textbf{0.066} & \textbf{0.061} & 0.048 & \textbf{0.035}\\ \cmidrule{2-15}
            & \multirow{2}{*}{{DReG}} & vanilla & 0.234 & 0.132 & 0.077 & 0.176 & 0.191 & 0.088 & 0.153 & 0.094 & 0.096 & 0.085 & 0.050 & 0.037\\
            & & \ours & 0.168 & \textbf{0.075} & \textbf{0.065} & \textbf{0.139} & 0.103 & \textbf{0.055} & \textbf{0.042} & \textbf{0.026} & \textbf{0.076} & \textbf{0.069} & 0.046 & 0.036\\ \midrule 
            & \multirow{2}{*}{{HI-VAE}} & vanilla & 0.127 & 0.107 & 0.126 & 0.114 & 0.141 & 0.079 & 0.105 & 0.044 & 0.100 & 0.098 & 0.062 & 0.039 \\
            & & \ours & \textbf{0.081} & 0.060 & \textbf{0.117} & \textbf{0.011} & \textbf{0.095} & 0.049 & 0.109 & \textbf{0.024} & \textbf{0.069} & \textbf{0.015} & \textbf{0.033} & 0.038\\  
            \bottomrule
        \end{tabular}
    }
\end{table*}
}

\subsection{Heterogeneous Data}\label{subsec:heterogeneous}

{

\sisetup{table-format=1.3, round-mode = places, round-precision = 3, detect-all}
\begin{table}[t]
    \centering
    \caption{Error on the heterogeneous experiments for the baseline and our framework, aggregated by type of likelihood.} \label{tab:types}
    {
        \begin{tabular}{l|SSSS} 
            \toprule 
            & {Poisson} & {Cat.} & {$\log\mathcal{N}$} & {$\mathcal{N}$} \\ \midrule
            vanilla & \bfseries 0.058283 & 0.157897 & 0.064481 & 0.040884 \\
            \ours & 0.082903 & \bfseries 0.065499 & \bfseries 0.056633 & \bfseries 0.038784 \\
            \bottomrule
        \end{tabular}
    }
\end{table}

}

\begin{figure}[t]
    \centering
    \includegraphics[width=\linewidth, keepaspectratio]{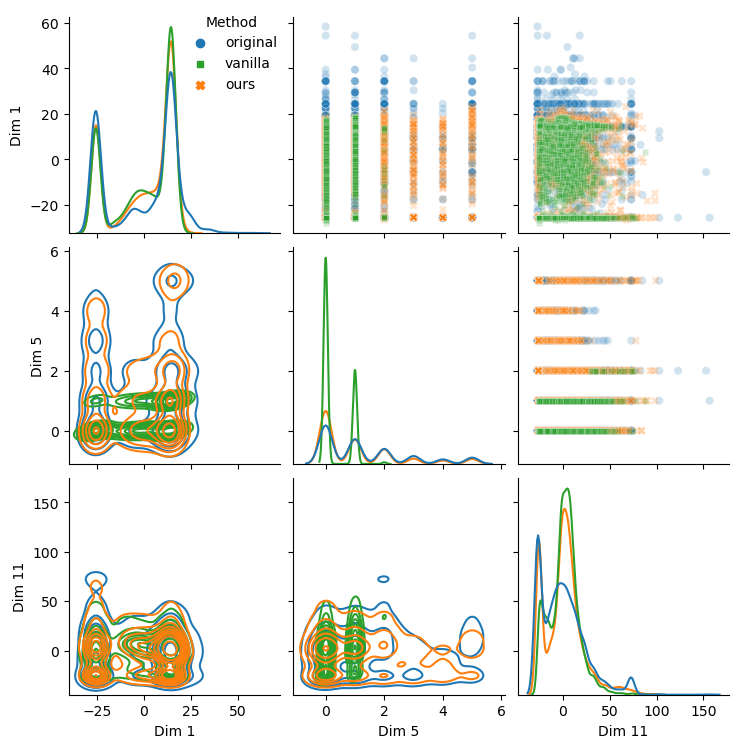}
    \caption{Pair plot of 3 dimensions of \textit{HI}, generated from different VAE models. Diagonal show the marginals, upper-diagonals scatter plots, and lower-diagonals kernel density estimates. The VAE trained with our approach is able to generate faithful samples.}
    \label{fig:pairplot}
\end{figure}

We first turn our attention to heterogeneous data modeling. 
While the task may look simple at first, %
we need to deal with plenty of modalities, each one with unique properties. 
Moreover, models are comparatively simple, forming a breeding ground for \problem.

We use as models VAEs as the one introduced in \cref{sec:mm-and-optim}, using as objective the ELBO~\citep{diederik2014auto}, IWAE~\citep{burda2016importance}, and DReG~\cite{tucker2018dreg}.
Additionally, we include HI-VAE~\citep{nazabal2020hivae} as an example of tailored heterogeneous model (see \cref{sec:hivae}).

We consider 12 datasets collected from the UCI~\citep{Dua:2019} and R~\citep{Rpackage} repositories, covering a wide range of dataset sizes and likelihoods.
We assign 4 likelihood types (normal, log-normal, Poisson, and categorical) depending on the modality domain.
Since likelihoods are not comparable, we use as metric the normalized mean squared error (for numerical data) and error rate (for categorical data), similar to~\citet{nazabal2020hivae}.

\textbf{Do we reconstruct better?} 
Explaining the observed data explicitly appears in the objective function (\cref{eq:iwae}).
If our approach works, reconstruction error should be reduced as a result of impartialy learning to explain all modalities.
\Cref{tab:test_errors} (left) shows the reconstruction error for 9 heterogeneous datasets, for which the models trained with our approach improve over the vanilla case in a statistically significant manner in 30 out of 36 cases.
Interestingly, our approach specially benefits the standard heterogenous VAE model, outperforming the HI-VAE (trained with both vanilla and impartial optimization) in several datasets. Importantly, for the majority of datasets, the performance of  HI-VAE is significantly improved by impartial optimization, outperforming the rest of VAE models, e.g., in  \textit{Adult} and \textit{Diamonds}. %

\textbf{Where does the improvement come from?}
We investigate whether any likelihood type benefits from our framework.
\cref{tab:types} shows again reconstruction error, this time aggregated by data type.  Here we can observe that we improve over all data types---and specially in categorical variables---by slightly worsening reconstruction on Poisson likelihoods.
In \cref{app:types}, we argue that the gradients of Poisson likelihoods are comparatively big among likelihood types, and thus dominates the learning process under standard optimization. 
Essentially, the trade-off found by our framework in \cref{tab:types} is the result of preventing this dominance.

\textbf{Does impartial optimization help in homogeneous settings?}
It is reasonable to suspect that \problem only appears when each modality uses a different likelihood type.
Assigning now exclusively normal likelihoods, we show in \cref{tab:test_errors} (right) that \problem  also occurs in homogeneous settings, and that our approach may  significantly improve model training even if all modalities share the same data type.

\textbf{Can we generate faithful data?} 
A key aspect of heterogeneous modeling is data generation. 
As a qualitative example, we train on the \textit{HI} dataset a VAE-ELBO, using vanilla and impartial optimization. 
We show  generated samples by the two VAEs in \cref{fig:pairplot} for three dimensions of the dataset, compared against the test data.
While both models similarly reconstruct the two continuous marginals, %
only we properly generate the categorical variable (middle), 
which concurs with the previous analysis.
More importantly, the VAE model trained with our framework is able to faithfully recreate the dependencies between modalities, as it can be observed in the off-diagonal figures.

\subsection{Multimodal Data}

We focus now on mixture-based multimodal VAE models. %
Besides the obvious architectural differences, these experiments are significantly more demanding and complex, involving millions of parameters and high-dimensional modalities.
We use \ac{SIWAE} (\cref{eq:tight-loss}) for most results in the main paper, as it is specially prone to \problem. %

We reproduce the setups of \citet{sutter2021mopoe} and \citet{shi2019mmvae},
using the same architectures, and taking as dataset {MNIST-SVHN-Text}, which randomly matches positive pairs from {MNIST}~\citep{lecun2010mnist} and {SVHN}~\citep{netzer2011svhn}, and generates a one-hot-encoded text representing the label in common.
This is a well-suited dataset for our purposes, since the high disparity in number of dimensions should ease \problem during training.
\added{Note that in all experiments we divide the log-likelihood by the number of dimensions (local step, see \cref{sec:dealing-with}), to offer fair comparisons, as it is a common practice in the field.}

We consider MVAE, MMVAE, and MoPoE as models, which differ in the choice of experts ($\mathcal{A}\subset\mathcal{P}(D)$) for the posterior approximation, %
as explained in \cref{sec:mixture-based}.

\textbf{Do we reconstruct better?}
As a sanity check, we again check how well we are able to reconstruct each modality. 
Following the existing literature, we measure reconstruction capabilities in terms of \textit{generative coherence}.
Specifically, we generate latent samples using all the modalities as input, and reconstruct each modality $\rvx_d$. 
Then, we feed each of these %
samples into modality-specific digit classifiers, and compute the accuracy \wrt the ground-truth digit.
\Cref{tab:multimodal-reconstruction} shows that our framework improves reconstruction coherence for all cases and models, sometimes by a statistically significant margin. %
It is also worth-noting that, in the case of MoPoE, the statistical test is inconclusive as the vanilla case has large variances. %

{

\sisetup{table-format=2.2, round-mode = places, round-precision = 2, detect-all}
\begin{table}[t]
    \centering
    \caption{Reconstruction coherence ($A = \{M, S, T\}$) for each modality and model, trained using SIWAE.}
    \label{tab:multimodal-reconstruction}
    {
        \begin{tabular}{c@{\hspace{.5\tabcolsep}}l|SSS} 
            \toprule 
            & $\rvx_d$ & {M} & {S} & {T} \\ \midrule
            \multirow{2}{*}{MVAE} & vanilla & 97.37306952476501 & 87.47473120689392 & 98.82936596870422 \\
            & \ours & 97.41689711809158 & 87.62706816196442 & \bfseries 99.19884651899338 \\ \midrule
            \multirow{2}{*}{MMVAE} & vanilla & 58.94593670964241 & 61.26817092299461 & 63.27350363135338 \\
            & \ours & \bfseries 74.15943145751953 & 68.9321979880333 & \bfseries 78.16554307937622 \\ \midrule
            \multirow{2}{*}{MoPoE} & vanilla & 75.09585581719875 & 67.16080456972122 & 76.60753987729549 \\
            & \ours & 96.91308736801147 & 89.01377469301224 & 99.28467273712158 \\
            \bottomrule
        \end{tabular}
    }
\end{table}

}

{

\sisetup{table-format=2.2, round-mode = places, round-precision = 2, detect-all}
\begin{table*}[t]
    \centering
    \caption{Self and cross generation coherence (\%) results for different models on \textbf{M}NIST-\textbf{S}VHN-\textbf{T}ext, trained using \ac{SIWAE} and averaged over 5 different seeds. Models trained with our framework are able to sample more coherent modalities.} \label{tab:multimodal-siwae}
    \resizebox{\textwidth}{!}
    {
        \begin{tabular}{c@{\hspace{.5\tabcolsep}}l|SSS|SSS|SSS|SSS} 
            \toprule 
            & & \multicolumn{3}{c}{Self coherence} & \multicolumn{9}{c}{Cross coherence} \\
            & $\vx_d$ & {M} & {S} & {T} & \multicolumn{3}{c}{M} & \multicolumn{3}{c}{S} & \multicolumn{3}{c}{T} \\
            & $A$ & {M} & {S} & {T} & {S} & {T} & \multicolumn{1}{c|}{S,T} & {M} & {T} & \multicolumn{1}{c|}{M,T} & {M} & {S} & \multicolumn{1}{c}{M,S} \\ \midrule
            \multirow{2}{*}{MVAE} & vanilla & 82.06321358680725 & 12.078495621681211 & 36.668736934661866 & 10.342429280281062 & 17.118748845532533 & 19.186700284481048 & 49.992209672927856 & 19.312240332365033 & 31.188021302223207 & 62.50251233577728 & 10.815181583166118 & 64.25450205802917 \\
            & \ours & 87.63102412223815 & 12.466459572315217 & \bfseries 78.87604594230652 & 10.750340372323983 & \bfseries 25.992168784141534 & \bfseries 27.846901714801786 & 50.01845777034759 & \bfseries 33.13292294740677 & 29.62448984384537 & 61.17406725883484 & 11.669214963912959 & 63.63236486911774 \\\midrule
            \multirow{2}{*}{MMVAE} & vanilla & \bfseries 95.89911252260208 & 48.302508518099785 & 53.02141793072224 & 28.43314968049526 & 52.51514203846455 & 40.45413397252559 & 84.44346040487289 & 51.07617117464541 & 67.76809990406036 & \bfseries 96.80366516113281 & 39.96035195887089 & 68.38035881519318 \\ 
            & \ours & 95.89718679587045 & 58.19725841283798 & \bfseries 88.69847655296326 & \bfseries 49.32506904006004 & \bfseries 79.31589980920157 & \bfseries 64.30353770653406 & \bfseries 87.28511532147726 & \bfseries 76.17471218109131 & \bfseries 81.7105770111084 & 96.69901430606842 & 57.86134203275045 & 77.27593878904977 \\ \midrule
            \multirow{2}{*}{MoPoE} & vanilla & 92.31970608234406 & 11.597061343491073 & 69.05172914266586 & 10.13441793620586 & 51.02324187755585 & 34.667880460619926 & 41.930220276117325 & 46.385733783245087 & 51.58409625291824 & 85.18766462802887 & 10.566544160246848 & 67.53952503204346 \\
             & \ours & 90.99198579788208 & 11.996489390730854 & \bfseries 83.81824642419815 & \bfseries 10.6312271207571 & 62.75309696793556 & 52.08384543657303 & 28.190246224403376 & 46.90524935722351 & 43.34438294172287 & 79.64047342538834 & 10.81472355872392 & 90.32563269138336 \\
            \bottomrule
        \end{tabular}
    }
\end{table*}

}

\textbf{Do we improve conditional generation?} 
One desirable property of a multimodal model is generating coherent samples based on another modalities. 
In our case, this translates to generating samples of the same digit as the input.
We use again generative coherence as metric. 
This time, given an expert $A\subset \mathcal{P}(D)$, and an output modality $\vx_d$, we %
impute $\rvx_d \sim p_{\theta_d}(\rvx_d | \rmZ_A)$ and check if the imputed value matches the original digit.
\replaced{%
Besides, we distinguish between self coherence ($A = \{d\}$), and cross coherence (the average of all $A\in\mathcal{P}(D)$ such that $d\notin A$).
}{%
Besides, for each modality $d$ we distinguish between self coherence, where we compute the average accuracy of samples conditioned on that same modality ($A = \{d\}$); and cross coherence, where samples are instead conditioned on experts not containing that modality (every $A\in\mathcal{P}(D)$ such that $d\notin A$). %
}

\begin{figure}
    \hspace{-0.25cm}\includegraphics[keepaspectratio, width=\linewidth]{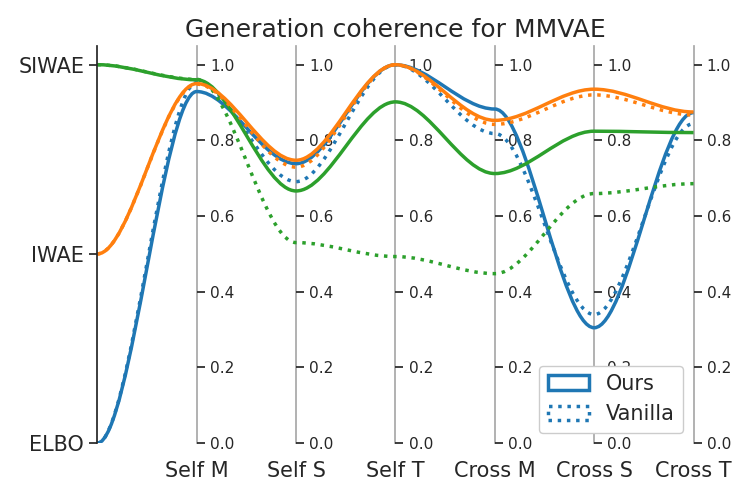} 
    \caption{Generation coherence of MMVAE with ELBO, IWAE, and SIWAE. We improve most metrics \wrt the baseline.}
    \label{fig:mmvae-parallel}
\end{figure}

\Cref{tab:multimodal-siwae} shows the self  and cross coherence results for all models and both approaches, trained with \ac{SIWAE}.
While there are trade-offs, we can observe that our framework in general improves both self  and cross coherence across all models.
For example, Text (T) and SVHN (S) were overlooked in MVAE and MMVAE, respectively, and the impartial VAE model increases self coherence for those modalities, as well as cross coherence when they appear in the expert $A$.
As mentioned in \cref{sec:mixture-based}, SIWAE is prone to \problem.
However, all objectives benefit from our framework.
\Cref{fig:mmvae-parallel} shows a parallel coordinate plot with the generative coherence results for MMVAE, evaluated on all objectives.
While SIWAE significantly improves with impartial optimization (as expected), we also improve all the different metrics for all losses.

\textbf{Do we generate more informative latent spaces?}
One key aspect of latent space models is that the latent space should be rich and informative. 
Following the existing literature, we evaluate the quality of a latent space by training a linear classifier to predict the ground-truth label, taking samples of $\rmZ$ as input.

Another key aspect, this time of mixture-based multimodal VAE models, is that the encoders should be as similar as possible (\ac{EEI}), and thus their latent spaces.
Just as before, here we distinguish between self and cross latent classification accuracy.
For each expert $A$, self latent classification refers to classifying test samples from the same expert the classifier was trained with, while cross latent classification refers to classifying test samples coming from an expert different from the one the classifier was trained with.

{

\sisetup{table-format=2.2, round-mode = places, round-precision = 2, detect-all}
\begin{table}[t]
    \centering
    \caption{Self and cross latent classification accuracy (\%) for different models and losses on MNIST-SVHN-Text.} \label{tab:multimodal-latent}
    {
        \begin{tabular}{c@{\hspace{.5\tabcolsep}}l|SSS} 
            \toprule 
            & & {ELBO} & {IWAE} & {SIWAE} \\ \midrule
            & & \multicolumn{3}{c}{Self latent classification} \\ \midrule
            \multirow{2}{*}{MVAE} & vanilla & 69.68467712402343 & 69.14131691058477 & 68.58165929714838 \\
            & \ours & 69.95231062173843 & 69.06403303146362 & \bfseries 69.75104610125223 \\ \midrule
            \multirow{2}{*}{MMVAE} & vanilla & 71.8134934703509 & 87.55302156011264 & 71.29535588125387 \\
            & \ours & \bfseries 87.82669926683107 & 90.78451957967548 & \bfseries 85.553377866745 \\ \midrule
            \multirow{2}{*}{MoPoE} & vanilla & 89.85391904910406 & 87.2267464796702 & 67.58138493945202 \\
            & \ours & \bfseries 91.46597236394882 & \bfseries 90.74468413988749 & \bfseries 69.26101893186569 \\ \midrule
            & & \multicolumn{3}{c}{Cross latent classification} \\ \midrule
            \multirow{2}{*}{MVAE} & vanilla & 33.59951532549328 & 39.15408961474895 & 38.362453712357414 \\
            & \ours & 35.24536229670047 & \bfseries 49.72874805745152 & \bfseries 46.22597529863317 \\ \midrule
            \multirow{2}{*}{MMVAE} & vanilla & 44.247944218417007 & 76.81059344775147 & 40.601979868693483 \\
            & \ours & \bfseries 71.42414665884442 & \bfseries 84.799057300444 & \bfseries 60.49764014228627 \\ \midrule
            \multirow{2}{*}{MoPoE} & vanilla & 66.13565652320782 & 83.7097362925609 & 40.36178290843963 \\
            & \ours & \bfseries 84.52272578659985 & \bfseries 90.47804905308618 & \bfseries 53.23550390700499 \\
            \bottomrule
        \end{tabular}
    }
\end{table}

}

We show in \cref{tab:multimodal-latent} the classification accuracies, averaged over experts.
We can observe that MMVAE and MoPoE significantly improve self latent classification accuracy when they are trained with our framework.
More importantly, all models significantly improve the cross latent classification accuracy, independently of the loss they were trained with, indicating that the latent spaces between experts are more similar between them (i.e., satisfy \ac{EEI}).

\setlength\tabcolsep{3pt} %

\textbf{Does impartial optimization add a lot of overhead?}
\begin{wraptable}[7]{r}{.5\linewidth}
    \centering
    \vspace{-.5\baselineskip}
    \begin{tabular}{CCC|CC} \toprule
         \text{\ac{LI}} & \text{\ac{EEI}} & \text{\ac{DEI}} & \text{time (h)} & \text{\#} \\ \midrule
         \openbox & \openbox & \openbox  & 10.06 & 0 \\ 
         \checkbox & \openbox & \openbox & 11.42 & D \\
         \checkbox & \checkbox & \openbox & 11.64 & 2D \\
         \checkbox & \checkbox & \checkbox & 11.89  & 3D \\
         \bottomrule
    \end{tabular}
\end{wraptable}
\added{The inset table} shows the training times for MMVAE as we change the number of blocks for which we apply \cref{alg:backward}.
As expected, the training time increases as we apply more MTL algorithms to the training.
In the case of MMVAE, we have 9 different \block{s}, 
and yet the training time increases only an \SI{18}{\percent}, going from \SI{10}{\hour} of training to \SI{11.89}{\hour}.
Each additional step increased in \SI{25}{\minute} the training time, which makes us believe that the extra overhead in the first transition is due to our implementation to manipulate the backward pass for \cref{alg:backward}.
\section{Conclusions\pages{0.25}}

In this work, we have studied the problem of \problem in multimodal VAEs, showing that it can be understood as a consequence of the conflict between gradients of different modalities during training.
We confined this conflict to a sub-graph of the computational graph, the \block, and proposed a general pipeline to enforce impartial optimization across modalities. %
We  have analyzed different tailored models, where several \block{s} may appear, %
proving the flexibility of our modular approach.
Finally, we have empirically shown that our approach can significantly improve the performance of these models on a range of datasets, losses and metrics.%

\added{We believe} this work opens venues for future \replaced{work}{research}.  First, as our method relies on off-the-shelf solutions from MTL,  it would be interesting to  develop  gradient-conflict solutions for the specifics of multimodal VAEs.
Second, exploring variations of \block{s} for specific applications, \eg, non-modular designs that reduce the current overhead, or \block{s} that take into account missing patterns in real-world data, could lead to exciting future works.

\section{Acknowledgements}

We would like to thank Pablo S\'anchez-Mart\'in for providing useful feedback on the manuscript, as well as to the anonymous reviewers and meta-reviewer who helped to improve the quality of the paper during the review process.

\bibliography{references}

\begin{thebibliography}{37}
\providecommand{\natexlab}[1]{#1}
\providecommand{\url}[1]{\texttt{#1}}
\expandafter\ifx\csname urlstyle\endcsname\relax
  \providecommand{\doi}[1]{doi: #1}\else
  \providecommand{\doi}{doi: \begingroup \urlstyle{rm}\Url}\fi

\bibitem[Baltru{\v{s}}aitis et~al.(2018)Baltru{\v{s}}aitis, Ahuja, and
  Morency]{baltruvsaitis2018multimodal}
Baltru{\v{s}}aitis, T., Ahuja, C., and Morency, L.-P.
\newblock Multimodal machine learning: A survey and taxonomy.
\newblock \emph{IEEE transactions on pattern analysis and machine
  intelligence}, 41\penalty0 (2):\penalty0 423--443, 2018.

\bibitem[Barrej{\'o}n et~al.(2021)Barrej{\'o}n, Olmos, and
  Art{\'e}s-Rodr{\'\i}guez]{shivae}
Barrej{\'o}n, D., Olmos, P.~M., and Art{\'e}s-Rodr{\'\i}guez, A.
\newblock Medical data wrangling with sequential variational autoencoders.
\newblock \emph{arXiv preprint arXiv:2103.07206}, 2021.
\newblock URL \url{https://arxiv.org/abs/2103.07206}.

\bibitem[Burda et~al.(2016)Burda, Grosse, and
  Salakhutdinov]{burda2016importance}
Burda, Y., Grosse, R.~B., and Salakhutdinov, R.
\newblock Importance weighted autoencoders.
\newblock In Bengio, Y. and LeCun, Y. (eds.), \emph{4th International
  Conference on Learning Representations, {ICLR} 2016, San Juan, Puerto Rico,
  May 2-4, 2016, Conference Track Proceedings}, 2016.
\newblock URL \url{http://arxiv.org/abs/1509.00519}.

\bibitem[Chen et~al.(2018)Chen, Badrinarayanan, Lee, and
  Rabinovich]{chen2017gradnorm}
Chen, Z., Badrinarayanan, V., Lee, C., and Rabinovich, A.
\newblock Gradnorm: Gradient normalization for adaptive loss balancing in deep
  multitask networks.
\newblock In Dy, J.~G. and Krause, A. (eds.), \emph{Proceedings of the 35th
  International Conference on Machine Learning, {ICML} 2018,
  Stockholmsm{\"{a}}ssan, Stockholm, Sweden, July 10-15, 2018}, volume~80 of
  \emph{Proceedings of Machine Learning Research}, pp.\  793--802. {PMLR},
  2018.
\newblock URL \url{http://proceedings.mlr.press/v80/chen18a.html}.

\bibitem[Chen et~al.(2020)Chen, Ngiam, Huang, Luong, Kretzschmar, Chai, and
  Anguelov]{graddrop}
Chen, Z., Ngiam, J., Huang, Y., Luong, T., Kretzschmar, H., Chai, Y., and
  Anguelov, D.
\newblock Just pick a sign: Optimizing deep multitask models with gradient sign
  dropout.
\newblock In Larochelle, H., Ranzato, M., Hadsell, R., Balcan, M., and Lin, H.
  (eds.), \emph{Advances in Neural Information Processing Systems 33: Annual
  Conference on Neural Information Processing Systems 2020, NeurIPS 2020,
  December 6-12, 2020, virtual}, 2020.
\newblock URL
  \url{https://proceedings.neurips.cc/paper/2020/hash/16002f7a455a94aa4e91cc34ebdb9f2d-Abstract.html}.

\bibitem[Chennupati et~al.(2019)Chennupati, Sistu, Yogamani, and
  Rawashdeh]{Chennupati2019MultiNetMF}
Chennupati, S., Sistu, G., Yogamani, S.~K., and Rawashdeh, S.~A.
\newblock Multinet++: Multi-stream feature aggregation and geometric loss
  strategy for multi-task learning.
\newblock \emph{2019 IEEE/CVF Conference on Computer Vision and Pattern
  Recognition Workshops (CVPRW)}, pp.\  1200--1210, 2019.

\bibitem[Dua \& Graff(2017)Dua and Graff]{Dua:2019}
Dua, D. and Graff, C.
\newblock {UCI} machine learning repository, 2017.
\newblock URL \url{http://archive.ics.uci.edu/ml}.

\bibitem[Ghosh et~al.(2020)Ghosh, Sajjadi, Vergari, Black, and
  Sch{\"{o}}lkopf]{ghosh2019variational}
Ghosh, P., Sajjadi, M. S.~M., Vergari, A., Black, M.~J., and Sch{\"{o}}lkopf,
  B.
\newblock From variational to deterministic autoencoders.
\newblock In \emph{8th International Conference on Learning Representations,
  {ICLR} 2020, Addis Ababa, Ethiopia, April 26-30, 2020}. OpenReview.net, 2020.
\newblock URL \url{https://openreview.net/forum?id=S1g7tpEYDS}.

\bibitem[Guo et~al.(2019)Guo, Wang, and Wang]{guo2019deep}
Guo, W., Wang, J., and Wang, S.
\newblock Deep multimodal representation learning: A survey.
\newblock \emph{IEEE Access}, 7:\penalty0 63373--63394, 2019.

\bibitem[Kendall et~al.(2018)Kendall, Gal, and Cipolla]{kendall2018multi}
Kendall, A., Gal, Y., and Cipolla, R.
\newblock Multi-task learning using uncertainty to weigh losses for scene
  geometry and semantics.
\newblock In \emph{2018 {IEEE} Conference on Computer Vision and Pattern
  Recognition, {CVPR} 2018, Salt Lake City, UT, USA, June 18-22, 2018}, pp.\
  7482--7491. {IEEE} Computer Society, 2018.
\newblock \doi{10.1109/CVPR.2018.00781}.
\newblock URL
  \url{http://openaccess.thecvf.com/content\_cvpr\_2018/html/Kendall\_Multi-Task\_Learning\_Using\_CVPR\_2018\_paper.html}.

\bibitem[Kingma \& Welling(2014)Kingma and Welling]{diederik2014auto}
Kingma, D.~P. and Welling, M.
\newblock Auto-encoding variational bayes.
\newblock In Bengio, Y. and LeCun, Y. (eds.), \emph{2nd International
  Conference on Learning Representations, {ICLR} 2014, Banff, AB, Canada, April
  14-16, 2014, Conference Track Proceedings}, 2014.
\newblock URL \url{http://arxiv.org/abs/1312.6114}.

\bibitem[LeCun et~al.(2010)LeCun, Cortes, and Burges]{lecun2010mnist}
LeCun, Y., Cortes, C., and Burges, C.
\newblock Mnist handwritten digit database.
\newblock \emph{ATT Labs [Online]}, 2, 2010.
\newblock URL \url{http://yann.lecun.com/exdb/mnist}.

\bibitem[Liu et~al.(2021{\natexlab{a}})Liu, Liu, Jin, Stone, and Liu]{cagrad}
Liu, B., Liu, X., Jin, X., Stone, P., and Liu, Q.
\newblock Conflict-averse gradient descent for multi-task learning.
\newblock \emph{Advances in Neural Information Processing Systems}, 34,
  2021{\natexlab{a}}.

\bibitem[Liu et~al.(2021{\natexlab{b}})Liu, Li, Kuang, Xue, Chen, Yang, Liao,
  and Zhang]{liu2021imtl}
Liu, L., Li, Y., Kuang, Z., Xue, J.-H., Chen, Y., Yang, W., Liao, Q., and
  Zhang, W.
\newblock Towards impartial multi-task learning.
\newblock In \emph{International Conference on Learning Representations},
  2021{\natexlab{b}}.
\newblock URL \url{https://openreview.net/forum?id=IMPnRXEWpvr}.

\bibitem[Ma et~al.(2020)Ma, Tschiatschek, Turner, Hern\'{a}ndez-Lobato, and
  Zhang]{NEURIPS2020vaem}
Ma, C., Tschiatschek, S., Turner, R., Hern\'{a}ndez-Lobato, J.~M., and Zhang,
  C.
\newblock Vaem: a deep generative model for heterogeneous mixed type data.
\newblock In Larochelle, H., Ranzato, M., Hadsell, R., Balcan, M.~F., and Lin,
  H. (eds.), \emph{Advances in Neural Information Processing Systems},
  volume~33, pp.\  11237--11247. Curran Associates, Inc., 2020.
\newblock URL
  \url{https://proceedings.neurips.cc/paper/2020/file/8171ac2c5544a5cb54ac0f38bf477af4-Paper.pdf}.

\bibitem[Mehrasa et~al.(2019)Mehrasa, Jyothi, Durand, He, Sigal, and
  Mori]{Mehrasa_2019_CVPR}
Mehrasa, N., Jyothi, A.~A., Durand, T., He, J., Sigal, L., and Mori, G.
\newblock A variational auto-encoder model for stochastic point processes.
\newblock In \emph{{IEEE} Conference on Computer Vision and Pattern
  Recognition, {CVPR} 2019, Long Beach, CA, USA, June 16-20, 2019}, pp.\
  3165--3174. Computer Vision Foundation / {IEEE}, 2019.
\newblock \doi{10.1109/CVPR.2019.00328}.
\newblock URL
  \url{http://openaccess.thecvf.com/content\_CVPR\_2019/html/Mehrasa\_A\_Variational\_Auto-Encoder\_Model\_for\_Stochastic\_Point\_Processes\_CVPR\_2019\_paper.html}.

\bibitem[Morningstar et~al.(2021)Morningstar, Vikram, Ham, Gallagher, and
  Dillon]{Morningstar2021selbo}
Morningstar, W.~R., Vikram, S.~M., Ham, C., Gallagher, A.~G., and Dillon, J.~V.
\newblock Automatic differentiation variational inference with mixtures.
\newblock In Banerjee, A. and Fukumizu, K. (eds.), \emph{The 24th International
  Conference on Artificial Intelligence and Statistics, {AISTATS} 2021, April
  13-15, 2021, Virtual Event}, volume 130 of \emph{Proceedings of Machine
  Learning Research}, pp.\  3250--3258. {PMLR}, 2021.
\newblock URL \url{http://proceedings.mlr.press/v130/morningstar21b.html}.

\bibitem[Nadeau \& Bengio(2003)Nadeau and Bengio]{corrected-t-test}
Nadeau, C. and Bengio, Y.
\newblock Inference for the generalization error.
\newblock \emph{Machine learning}, 52\penalty0 (3):\penalty0 239--281, 2003.

\bibitem[Nazabal et~al.(2020)Nazabal, Olmos, Ghahramani, and
  Valera]{nazabal2020hivae}
Nazabal, A., Olmos, P.~M., Ghahramani, Z., and Valera, I.
\newblock Handling incomplete heterogeneous data using vaes.
\newblock \emph{Pattern Recognition}, 107:\penalty0 107501, 2020.

\bibitem[Netzer et~al.(2011)Netzer, Wang, Coates, Bissacco, Wu, and
  Ng]{netzer2011svhn}
Netzer, Y., Wang, T., Coates, A., Bissacco, A., Wu, B., and Ng, A.~Y.
\newblock Reading digits in natural images with unsupervised feature learning.
\newblock \emph{NeurIPS Workshop on Deep Learning and Unsupervised Feature
  Learning}, 2011.

\bibitem[{R Core Team}(2021)]{Rpackage}
{R Core Team}.
\newblock \emph{R: A Language and Environment for Statistical Computing}.
\newblock R Foundation for Statistical Computing, Vienna, Austria, 2021.
\newblock URL \url{https://www.R-project.org/}.

\bibitem[Rainforth et~al.(2018)Rainforth, Kosiorek, Le, Maddison, Igl, Wood,
  and Teh]{rainforth2018tighter}
Rainforth, T., Kosiorek, A.~R., Le, T.~A., Maddison, C.~J., Igl, M., Wood, F.,
  and Teh, Y.~W.
\newblock Tighter variational bounds are not necessarily better.
\newblock In Dy, J.~G. and Krause, A. (eds.), \emph{Proceedings of the 35th
  International Conference on Machine Learning, {ICML} 2018,
  Stockholmsm{\"{a}}ssan, Stockholm, Sweden, July 10-15, 2018}, volume~80 of
  \emph{Proceedings of Machine Learning Research}, pp.\  4274--4282. {PMLR},
  2018.
\newblock URL \url{http://proceedings.mlr.press/v80/rainforth18b.html}.

\bibitem[Reddi et~al.(2018)Reddi, Kale, and Kumar]{j.2018on}
Reddi, S.~J., Kale, S., and Kumar, S.
\newblock On the convergence of adam and beyond.
\newblock In \emph{6th International Conference on Learning Representations,
  {ICLR} 2018, Vancouver, BC, Canada, April 30 - May 3, 2018, Conference Track
  Proceedings}. OpenReview.net, 2018.
\newblock URL \url{https://openreview.net/forum?id=ryQu7f-RZ}.

\bibitem[Roeder et~al.(2017)Roeder, Wu, and Duvenaud]{Roeder2017StickingTL}
Roeder, G., Wu, Y., and Duvenaud, D.
\newblock Sticking the landing: Simple, lower-variance gradient estimators for
  variational inference.
\newblock In Guyon, I., von Luxburg, U., Bengio, S., Wallach, H.~M., Fergus,
  R., Vishwanathan, S. V.~N., and Garnett, R. (eds.), \emph{Advances in Neural
  Information Processing Systems 30: Annual Conference on Neural Information
  Processing Systems 2017, December 4-9, 2017, Long Beach, CA, {USA}}, pp.\
  6925--6934, 2017.
\newblock URL
  \url{https://proceedings.neurips.cc/paper/2017/hash/e91068fff3d7fa1594dfdf3b4308433a-Abstract.html}.

\bibitem[Ruder(2017)]{ruder2017overview}
Ruder, S.
\newblock An overview of multi-task learning in deep neural networks.
\newblock \emph{CoRR}, abs/1706.05098, 2017.
\newblock URL \url{http://arxiv.org/abs/1706.05098}.

\bibitem[Rumelhart et~al.(1986)Rumelhart, Hinton, and
  Williams]{Rumelhart1986backpropagation}
Rumelhart, D.~E., Hinton, G.~E., and Williams, R.~J.
\newblock Learning representations by back-propagating errors.
\newblock \emph{Nature}, 323:\penalty0 533--536, 1986.

\bibitem[Sener \& Koltun(2018)Sener and Koltun]{mgdaub}
Sener, O. and Koltun, V.
\newblock Multi-task learning as multi-objective optimization.
\newblock In Bengio, S., Wallach, H.~M., Larochelle, H., Grauman, K.,
  Cesa{-}Bianchi, N., and Garnett, R. (eds.), \emph{Advances in Neural
  Information Processing Systems 31: Annual Conference on Neural Information
  Processing Systems 2018, NeurIPS 2018, December 3-8, 2018, Montr{\'{e}}al,
  Canada}, pp.\  525--536, 2018.
\newblock URL
  \url{https://proceedings.neurips.cc/paper/2018/hash/432aca3a1e345e339f35a30c8f65edce-Abstract.html}.

\bibitem[Shi et~al.(2019)Shi, Narayanaswamy, Paige, and Torr]{shi2019mmvae}
Shi, Y., Narayanaswamy, S., Paige, B., and Torr, P. H.~S.
\newblock Variational mixture-of-experts autoencoders for multi-modal deep
  generative models.
\newblock In Wallach, H.~M., Larochelle, H., Beygelzimer, A.,
  d'Alch{\'{e}}{-}Buc, F., Fox, E.~B., and Garnett, R. (eds.), \emph{Advances
  in Neural Information Processing Systems 32: Annual Conference on Neural
  Information Processing Systems 2019, NeurIPS 2019, December 8-14, 2019,
  Vancouver, BC, Canada}, pp.\  15692--15703, 2019.
\newblock URL
  \url{https://proceedings.neurips.cc/paper/2019/hash/0ae775a8cb3b499ad1fca944e6f5c836-Abstract.html}.

\bibitem[Shi et~al.(2021)Shi, Paige, Torr, and N]{shi2021relating}
Shi, Y., Paige, B., Torr, P., and N, S.
\newblock Relating by contrasting: A data-efficient framework for multimodal
  generative models.
\newblock In \emph{International Conference on Learning Representations}, 2021.
\newblock URL \url{https://openreview.net/forum?id=vhKe9UFbrJo}.

\bibitem[Srivastava et~al.(2014)Srivastava, Hinton, Krizhevsky, Sutskever, and
  Salakhutdinov]{JMLR:v15:dropout}
Srivastava, N., Hinton, G., Krizhevsky, A., Sutskever, I., and Salakhutdinov,
  R.
\newblock Dropout: A simple way to prevent neural networks from overfitting.
\newblock \emph{Journal of Machine Learning Research}, 15\penalty0
  (56):\penalty0 1929--1958, 2014.
\newblock URL \url{http://jmlr.org/papers/v15/srivastava14a.html}.

\bibitem[Sutter et~al.(2020)Sutter, Daunhawer, and Vogt]{sutter2020mmjsd}
Sutter, T.~M., Daunhawer, I., and Vogt, J.~E.
\newblock Multimodal generative learning utilizing jensen-shannon-divergence.
\newblock In \emph{NeurIPS}, 2020.

\bibitem[Sutter et~al.(2021)Sutter, Daunhawer, and Vogt]{sutter2021mopoe}
Sutter, T.~M., Daunhawer, I., and Vogt, J.~E.
\newblock Generalized multimodal {ELBO}.
\newblock In \emph{International Conference on Learning Representations}, 2021.
\newblock URL \url{https://openreview.net/forum?id=5Y21V0RDBV}.

\bibitem[Tucker et~al.(2019)Tucker, Lawson, Gu, and Maddison]{tucker2018dreg}
Tucker, G., Lawson, D., Gu, S., and Maddison, C.~J.
\newblock Doubly reparameterized gradient estimators for monte carlo
  objectives.
\newblock In \emph{7th International Conference on Learning Representations,
  {ICLR} 2019, New Orleans, LA, USA, May 6-9, 2019}. OpenReview.net, 2019.
\newblock URL \url{https://openreview.net/forum?id=HkG3e205K7}.

\bibitem[Vahdat \& Kautz(2020)Vahdat and Kautz]{nvidia-vae}
Vahdat, A. and Kautz, J.
\newblock Nvae: A deep hierarchical variational autoencoder.
\newblock In Larochelle, H., Ranzato, M., Hadsell, R., Balcan, M.~F., and Lin,
  H. (eds.), \emph{Advances in Neural Information Processing Systems},
  volume~33, pp.\  19667--19679. Curran Associates, Inc., 2020.
\newblock URL
  \url{https://proceedings.neurips.cc/paper/2020/file/e3b21256183cf7c2c7a66be163579d37-Paper.pdf}.

\bibitem[Wu \& Goodman(2018)Wu and Goodman]{Wu2018MVAE}
Wu, M. and Goodman, N.~D.
\newblock Multimodal generative models for scalable weakly-supervised learning.
\newblock In Bengio, S., Wallach, H.~M., Larochelle, H., Grauman, K.,
  Cesa{-}Bianchi, N., and Garnett, R. (eds.), \emph{Advances in Neural
  Information Processing Systems 31: Annual Conference on Neural Information
  Processing Systems 2018, NeurIPS 2018, December 3-8, 2018, Montr{\'{e}}al,
  Canada}, pp.\  5580--5590, 2018.
\newblock URL
  \url{https://proceedings.neurips.cc/paper/2018/hash/1102a326d5f7c9e04fc3c89d0ede88c9-Abstract.html}.

\bibitem[Xu et~al.(2017)Xu, Sun, Deng, and Tan]{xu2017variational-text}
Xu, W., Sun, H., Deng, C., and Tan, Y.
\newblock Variational autoencoder for semi-supervised text classification.
\newblock In Singh, S.~P. and Markovitch, S. (eds.), \emph{Proceedings of the
  Thirty-First {AAAI} Conference on Artificial Intelligence, February 4-9,
  2017, San Francisco, California, {USA}}, pp.\  3358--3364. {AAAI} Press,
  2017.
\newblock URL \url{http://aaai.org/ocs/index.php/AAAI/AAAI17/paper/view/14299}.

\bibitem[Yu et~al.(2020)Yu, Kumar, Gupta, Levine, Hausman, and
  Finn]{yu2020pcgrad}
Yu, T., Kumar, S., Gupta, A., Levine, S., Hausman, K., and Finn, C.
\newblock Gradient surgery for multi-task learning.
\newblock In Larochelle, H., Ranzato, M., Hadsell, R., Balcan, M.~F., and Lin,
  H. (eds.), \emph{Advances in Neural Information Processing Systems},
  volume~33, pp.\  5824--5836. Curran Associates, Inc., 2020.
\newblock URL
  \url{https://proceedings.neurips.cc/paper/2020/file/3fe78a8acf5fda99de95303940a2420c-Paper.pdf}.

\end{thebibliography}
\bibliographystyle{icml2022}

\newpage
\onecolumn
\appendix

	\clearpage
	\newpage

\section{Multitask learning and conflicting gradients} \label{sec:mtl-methods}

The goal of \acdef[MTL]{multitask learning} is to simultaneously solve a set of $D$ tasks. 
Suppose that all of them share the input data $\rmX$, but each task defines its own loss function $\loss_d$.
To amortize parameters across tasks, one common
choice is to have a shared backbone, $\vx\mapsto\vy$, parameterized by $\theta_{sh}$, and a set of task-specific heads, $\vy\mapsto\etab_d$, where $\etab_d$ is the prediction for its associated task.
In order to learn the parameters, a common approach is to minimize the sum of losses, $\sum_d \loss_d$.

One main assumption in MTL is that of \textit{task impartiality}, which assumes that all tasks are equally important to solve, \ie, we do not prefer learning one task over another~\citep{liu2021imtl}.
MTL often suffers from \textit{negative transfer}, which is defined as the negative effect that simultaneously learning some tasks can have on the final model performance~\citep{ruder2017overview}.

Akin to this work, one research direction in MTL studies conflicting gradients in order to explain the existence of negative transfer. 
Indeed, it is easy to observe that the gradient \wrt the shared parameters is of the form $\sum_d \nabla_{\theta_{sh}} \loss_d$, and thus gradient differences make the model lean toward prioritizing some tasks over others.

\subsection{Conflicting-gradient solutions}

As explained in \cref{sec:dealing-with}, we consider MTL solutions to conflicting gradients, $f_\psi$ that modify the gradients during the backward pass.
These solutions can be classified in two main categories.
\begin{itemize}
    \item On the one hand, we have algorithms $f_\psi$ that scale each gradient $\vg_d$ according to a specific criterion, in order to deal with the disparities of gradients due to their magnitudes. That is, they replace each gradient $\vg_d$ with $\omega_d\,\vg_d$, where each algorithm $f_\psi$ sets the value of $\omega_d$ in each step differently.
    \item Second, direction-aware algorithms. These algorithms attempt to solve issues related with gradients pointing towards different directions of the parameter space, thus cancelling out each other.
\end{itemize}

We consider for all our experiments the following existing algorithms from the MTL literature:
\begin{itemize}
    \item Magnitude-aware: 
    \begin{enumerate}[i)]
        \item GradNorm~\citep{chen2017gradnorm} (GN) - Scales the gradients and try to normalize the magnitude of the gradients over time. Moreover, a hyperparameter $\alpha$ controls the intensity for which to normalize these gradients, using the ratio between task losses as a measure of the task convergence.
        
        In this work we have slightly modified GradNorm, such that instead of using the task losses, we use the magnitude of the gradients as a criterion to identify the ``task'' convergence.
        
        \item MGDA-UB~\citep{mgdaub} - Scales the gradient by finding the convex sum of the gradients that results in the minimum norm, such that advancing in that direction reduces all task losses.
        
        \item IMTL-G~\citep{liu2021imtl} - Scales the gradients by optimizing the scaling factors via a closed-form solution, such that the aggregated gradient (sum of raw gradients weighted by the scaling factors) has equal projections onto individual tasks.
        
        \item CAGrad~\citep{cagrad} - Generalization of MGDA-UB that introduces a hyperparameter $\alpha$ to control how much the resulting gradient direction differs from the one followed by SGD. 
    \end{enumerate}
    \item Direction-aware:
    \begin{enumerate}[i)]
        \item GradDrop~\citep{graddrop} (GD) - Randomly drops elements of the task gradients based on how much they conflict in direction with the aggregated gradient, such that those directions ``self-correct'' themselves and align with the rest of gradients.
        
        \item PCGrad~\citep{yu2020pcgrad} (PG) - Randomly projects task gradients between them, thus removing the orthogonal parts that would cancel out when computing the aggregated gradient.
    \end{enumerate}
\end{itemize}

For the heterogeneous experiments, we find the best $f_\psi$ by combining magnitude-aware solutions followed by direction-aware solutions, since they are cheaper to compute, and we can run more experiments.
For the multimodal experiments, we do not consider combinations of algorithms, but the algorithms by themselves.

\section{Alleviating \problem} \label{app:tailored-models}

\begin{figure}[t]
    \centering
    \hfill %
    \begin{subfigure}[c]{.4\textwidth}
        \hspace{3em} \includestandalone[width=\linewidth, mode=image|tex]{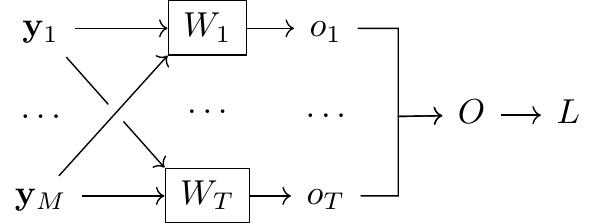}
        \caption{}
    \end{subfigure} %
    \hfill %
    \begin{subfigure}[c]{.55\textwidth}
        \includestandalone[width=\linewidth, mode=image|tex]{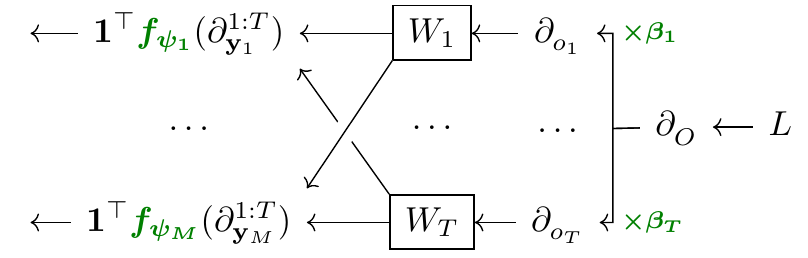}
        \caption{}
    \end{subfigure} %
    \hfill %
    \caption{Generic sketches of a (a) forward and a (b) modified backward pass of an \block. Note that here we are using the shorthand $\spd[1:K]{x}$ for a sequence of $K$ gradients \wrt $x$ (not to be confused with the partial derivative).}
    \label{fig:ours-algorithm}
\end{figure}

\begin{algorithm}[tb]
	\caption{Generic impartial backward pass within the \block.}
	\label{alg:backward-full}
	\begin{algorithmic}[1]
	    \STATE {\bfseries Definition} \textsc{ImpartialBackward}(inputs: $\vy_{1:M}$; heads: $W_{1:T}$; output: $O$)
		\STATE {\bfseries Input:} Output gradient, $\spd{O}$.
		\FOR{$t=1$ {\bfseries to} $T$}
		\STATE $\spd{o_t} \gets \textcolor{green!50!black}{\bm {\beta_t}}  \nabla_{o_t} O\, \spd{O}$ \hfill $\triangleright$ Re-weigh the gradient at the heads.
		\STATE $\nabla_{W_t} \loss \gets \nabla_{W_t} o_t\, \cdot \spd{o_t}$ \hfill $\triangleright$ Gradient of the head parameters (if any).
		    \FOR{$m=1$ {\bfseries to} $M$}
		        \STATE $\spd[t]{\vy_m} \gets \nabla_{\vy_m} o_t\, \cdot \spd{o_t}$ \hfill $\triangleright$ Per-task gradients \wrt the common inputs.
		    \ENDFOR
		\ENDFOR
		\FOR{$m=1$ {\bfseries to} $M$}
            \STATE {\bfseries backpropagate} $\bm{1}^\top \textcolor{green!50!black}{\bm {f_{\psi_m}}}(\spd[1:T]{\vy_m})$ {\bfseries through} $\vy_m$  \hfill $\triangleright$ Apply MTL methods and backpropagate the changes.
	    \ENDFOR
	\end{algorithmic}
\end{algorithm}

In \cref{sec:mm-and-optim} of the main paper, we have introduced the \block gradually, starting with a simple example, and showing how to adapt it as we were facing different challenges.
Here, we introduce the \block in a generic and flexible way, so that it could be easier for the reader to understand how to apply it to the tailored models explained in the main manuscript, as well as how to use the \block for their own use-cases.

\begin{figure}[t]
    \centering
    \hfill %
    \begin{subfigure}[c]{.4\textwidth}
        \includestandalone[width=\linewidth, mode=image|tex]{figs/mmvae}
        \caption{$\rmZ_d$ perspective.}
    \end{subfigure} %
    \hfill %
    \begin{subfigure}[c]{.55\textwidth}
        \includestandalone[width=\linewidth, mode=image|tex]{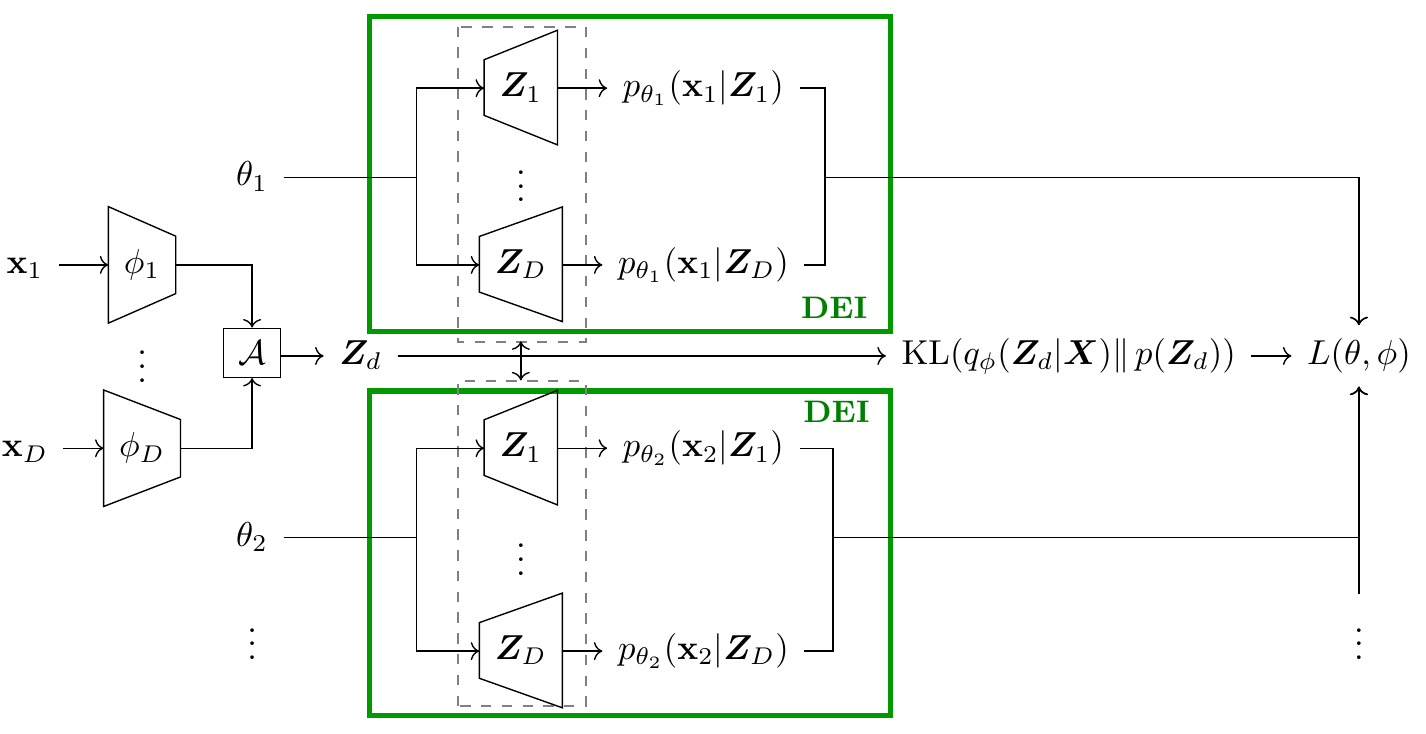}
        \caption{$\theta_d$ perspective.}
    \end{subfigure} %
    \hfill %
    \caption{Forward pass of MMVAE from two different points of view: as in~\cref{sec:mm-and-optim}, (a) shows the perspective where $\rmZ$ is the active role, where for each $\rmZ_A$ we find two \block{s}; in (b) we show the perspective of the decoder parameters, where the decoder parameters play the active role, and now we can observe that there is an \block result of evaluating each decoder in each expert samples. Here, the dashed block simply indicates that the sample $\rmZ_d$ gets distributed into that set of blocks to be evaluated.
    {Note that we explicitly show only 2 out of $D$ DEI blocks.}
    }
    \label{fig:mmvae-theta}
\end{figure}

\Cref{alg:backward-full} shows the new algorithm, and \cref{fig:ours-algorithm} the forward and backward pass. To detach the block from its original presentation, we have adopted here a generic notation for the different elements of the block, as well as allow for multiple entries. 
In this way, we would like to emphasize that the key aspect of the \block is its structure, and not the variables that appear within it. 
In other words, \cref{alg:backward-full} can be applied to any \block, independently of whether the input is an intermediate feature (such as in the blocks related with \ac{LI}, see \cref{sec:mm-and-optim}), or the features of a neural network (such as in the blocks related with \ac{DEI}, see \cref{sec:mixture-based-conflict}).
As for the last example, we could not show a computational block for the mixture-base models that introduces the three types of \block{s} at once. To help the reader, we present in \cref{fig:mmvae-theta} two different ways of drawing the computational graph of mixture-based models that unveil all the \block{s}.

With the re-formulation of the \block, we provide here a summary of the \block{s} presented in the models of the main paper:

\newcommand{\blockfunction}{\textsc{ImpartialBackward}}

\begin{center}
    \begin{tabular}{lcCC} \toprule
        Model & Goal & {\#} & \text{Backward call} \\ \midrule
         VAE & \ac*{LI} & 1 & \blockfunction(\vy; \omega_{1:D}; \etab) \\
         IWAE & \ac*{LI} & 1 & \blockfunction(\vy; \omega_{1:D}; \etab) \\
         DReG & \ac*{LI} & 2 & \blockfunction(\vy; \omega_{1:D}; \etab) \\
         HI-VAE & \ac*{LI} & 1 & \blockfunction([\vy, \mS]; \omega_{1:D}; \etab) \\ \midrule
         mixture-based & \ac*{LI} & |\mathcal{A}| & \blockfunction(\rmZ_A; \theta_{1:D}; p_\theta(\rmX | \rmZ_A)) \\ 
         mixture-based & \ac*{EEI} & |\mathcal{A}| & \blockfunction(\rmZ_A; \phi_\mathcal{A}; q_\phi(\rmZ_A | \rmX)) \\
         mixture-based & \ac*{DEI} & D & \blockfunction(\theta_d; \rmZ_\mathcal{A}; \loss) \\ 
         \bottomrule
    \end{tabular}
\end{center}

\section{Dominance of Poisson likelihoods} \label{app:types}

In this section, we attempt to mathematically sketch the results obtained in \cref{tab:types} of the main paper.
To do that, we are simply going to compute the expected value of the squared norm of the gradient with respect to each of the likelihoods, that is, we estimate $\Eop[\rvx_d]{\norm{\nabla_{\etab_d} \log p_{\theta_d}(\rvx_d; \etab_d)}^2}$.
We further simplify things by making the assumption that $\rvx_d$ actually follows the distribution $p_{\theta_d}(\rvx_d; \etab_d)$.
While unrealistic, this assumption should become more and more real as the training progresses.
We break down this informal proof in two steps:

\paragraph{Computing the expected squared norms.} 
We first take advantage that all considered distributions are part of the exponential distribution, and find a general formula valid for all of them.
As a reminder, the exponential family, with natural parameters $\etab\in\R^I$, is a family of distributions which is characterized by having a density function of the form
\begin{equation}
    p(\rvx; \etab) = h(\rvx) \exp\left( \bm T(\rvx)^\top \etab - A(\etab) \right),
\end{equation}
where each member of the family defines the values for: $h(\rvx)$, the base measure; $\bm T(\rvx)$ the sufficient statistics; and $A(\etab)$, the log-partition function.
Using this general expression, we can compute the value of $\Eop[\rvx_d]{\norm{\nabla_{\etab_d} \log p_{\theta_d}(\rvx_d; \etab_d)}^2}$:

\begin{align}
    \ln p(\rvx; \etab) &= \bm T(\rvx)^\top \etab - A(\etab) + C(\rvx), \\
    \spd{\eta_i} \ln p(\rvx; \etab) &= T_i(\rvx) - \spd{\eta_i} A(\etab) = T_i(\rvx) - \Eop{T_i(\rvx)}, \\
    \Eop[\rvx_d]{\norm{\nabla_{\etab_d} \log p_{\theta_d}(\rvx_d; \etab_d)}^2} &= \sum_i \Eop[\rvx_d]{\left( T_i(\rvx) - \Eop{T_i(\rvx)} \right)^2},
\end{align}
where we have used the fact that $\spd{\eta_i} A(\etab) = \Eop{T_i(\rvx)}$.

We can now simply plug in the specific values for the sufficient statistics for each of the likelihoods:
\begin{center}
    \begin{tabular}{l|CCC}
        & \spd{\eta_1} \ln p(\rvx; \etab) & \spd{\eta_2} \ln p(\rvx; \etab) & \Eop{\norm{\nabla_\etab \ln p(\rvx;\etab)}^2} \\ \midrule
        Normal & \rvx - \mu & \rvx^2 - (\mu^2 + \sigma^2) & \sigma^2 + 4\mu^2\sigma^2 \\
        Log-normal & \ln\rvx - \mu & (\ln\rvx)^2 - (\mu^2 + \sigma^2) & \sigma^2 + 4\mu^2\sigma^2 \\
        Poisson & \rvx - \lambda & & \lambda \\
        Categorical & [\rvx = i] - \pi_i & & \sum_{i=1}^I \Eop{([\rvx = i] - \pi_i)^2} 
    \end{tabular}
\end{center}

For each likelihood above, we have used the usual notation for their normal parameters. 
Moreover, notice that the moments are not well-defined for the categorical distribution. Instead, we just compute the average over the entire dataset. Here, $[\rvx = i]$ denotes the Iverson brackets (whether $\rvx_i$ pertains to the $i$-th class).

\paragraph{Bounding the norms under our working pipeline.} 
Once that we have rough estimates of the expected squared norms of the gradients for each likelihood, we need to come down to earth and connect it with the experiments in \cref{subsec:heterogeneous}.
Specifically, we need to take into account the preprocessing and the datasets themselves. We use the \textit{Adult} dataset as an example:
\begin{itemize}
    \item Normal: We standardize normal data, such that $\mu = 0$ and $\sigma = 1$. Therefore, $\Eop{\norm{\nabla_\etab \ln p(\rvx;\etab)}^2} \approx 1$.
    \item Log-normal: We standardize (without shifting) in log-space. In \textit{Adult}, the biggest log-normal distribution lies in the range $[15, 22]$, such that $\mu\approx 1$ and $\sigma<1$ in log-scale, and $\Eop{\norm{\nabla_\etab \ln p(\rvx;\etab)}^2} \approx 1$.
    \item Poisson: Since data is discrete, we do not standardize it. Count data can be quite large, reaching in \textit{Adult} a maximum value of $100$. Thus, $\Eop{\norm{\nabla_\etab \ln p(\rvx;\etab)}^2} >> 1$ in \textit{Adult}.
    \item Categorical: Again, we do not standardize categorical data, as it is discrete. However, it is relatively simple to see that $0\leq \Eop{\norm{\nabla_\etab \ln p(\rvx;\etab)}^2} \leq I$ since $0\leq\pi_i\leq1$ and $[\rvx=i]\in\{0,1\}$. However, the number of classes $I$ is usually small, and the gradient is bounded by $I$ during the entire training, while in the other cases they are not (we just considered the cases where we have the ground-truth parameters).
\end{itemize}

Therefore, using these rough calculations, we can expect the values of $\Eop{\norm{\nabla_\etab \ln p(\rvx;\etab)}^2}$ to lie in the following order:
\begin{center}
    Categorical $<$ Normal $\approx$ Log-normal $<<$ Poisson.
\end{center}
And, if we compute the difference between normalized errors in \cref{tab:types}, we obtain that our approach improves the error across types in an order similar to the reverse of the one shown above:
{

\sisetup{table-format=-1.3, round-mode = places, round-precision = 3, detect-all}
\begin{center}
    \begin{tabular}{l|SCSCSCS} 
        \toprule 
                & {Cat.} & & {$\log\mathcal{N}$} & & {$\mathcal{N}$}   & & {Poisson} \\ \midrule
        vanilla & 0.157897 & & 0.064481  & & 0.040884   & & \bfseries 0.058283 \\
        \ours    & \bfseries 0.065499    & & \bfseries 0.056633 & & \bfseries 0.038784& & 0.082903 \\ \midrule
        improvement & 0.09239800000000001  & > & 0.007847999999999994 & \approx & 0.0020999999999999977& > & -0.024620000000000003 \\
        \bottomrule
    \end{tabular}
\end{center}
    
}

\section{Model descriptions} \label{app:models}

{  %

\renewcommand{\rvz}{\rmZ}

\newcommand{\linear}[1]{{\small \texttt{[Linear-$#1$]}}}
\newcommand{\conv}[3]{{\small\texttt{[Conv-$#1$-$#2$-$#3$]}}}
\newcommand{\convt}[3]{{\small\texttt{[ConvT-$#1$-$#2$-$#3$]}}}
\newcommand{\dropout}[1]{{\small \texttt{[Dropout-\SI{#1}{\percent}]}}}
\newcommand{\batchnorm}{{\small \texttt{[BN]}}}
\renewcommand{\tanh}{{\small \texttt{[Tanh]}}}
\newcommand{\relu}{{\small \texttt{[ReLU]}}}
\newcommand{\Sigmoid}{{\small \texttt{[Sigmoid]}}}

In this section we explain the implementation details for each model, please refer to the original papers for a detailed explanation of each model.
We use the following notation to describe the models:
 
\begin{center}
    \begin{tabular}{ll}
        $D$ & Number of features. \\
        $D'$ & Total number of likelihood parameters. \\
        $\latentsize$ & Latent size. \\
        $h$ & Hidden size. \\
        \linear{h} & Linear layer with output of size $h$. \\
        \added{\conv{k}{s}{p}} & \added{Convolutional layer with kernel size $k$, stride $s$ and padding $p$.} \\
        \added{\convt{k}{s}{p}} & \added{Transposed convolutional layer with kernel size $k$, stride $s$ and padding $p$.} \\
        \dropout{10} & Dropout~\citet{JMLR:v15:dropout} with \SI{10}{\percent} of dropping probability. \\
        \relu & Rectified linear unit activation function. \\
        \tanh & Hyperbolic tangent activation function. \\
        \added{\Sigmoid} & \added{Sigmoid activation function.} \\
    \end{tabular}
\end{center}

\subsection{Variational autoencoder (VAE)}

We implement the original VAE~\citep{diederik2014auto} assuming the following probabilistic model:
\begin{center}
    \def\arraystretch{1.5}%
    \begin{tabular}{lL}
        Prior: & p(\rvz) = \mathcal{N}(0, I)  \\ 
        Likelihood: & p_\theta(\rmX | \rvz) = \prod_d p_d(\rvx_d | \eta_d(\rvz; \theta)) \\ 
        Variational approx.: & q_\phi(\rvz|\rmX) = \mathcal{N}(\mu(\rmX; \phi), \sigma(\rmX; \phi))
    \end{tabular}
\end{center}

Here $\mu$ and $\sigma$ are modelled by the encoder, and all $\eta_d$ are jointly modelled by the decoder. 

These two neural networks are of the following form:
\begin{center}
    \begin{tabular}{ll}
        \textbf{Encoder:} & \dropout{10}\batchnorm\linear{h}\tanh\linear{h}\tanh\linear{h}\tanh\linear{2\latentsize} \\
        \textbf{Decoder:} & \linear{h}\relu\linear{h}\relu\linear{h}\relu\linear{D'} \\
    \end{tabular}
\end{center}

Additionally, we make sure that each parameter fulfils its distributional constraints (e.g., the variance has to be positive) by passing it through a softplus function when necessary.
It is also important to note that, while we parametrize the latent space using the mean and standard deviation, we parametrize the parameters of the likelihoods using their natural parameters.

\paragraph{Loss.} We use the negative ELBO as training loss: 
\begin{equation}
    \operatorname{ELBO}(\rmX, \theta, \phi) \coloneqq \Eop[q_\phi]{\log p_\theta(\rmX|\rvz)} - \KLop{q_\phi(\rvz)}{p(\rvz)}.
\end{equation}

\paragraph{Imputation.} We impute data by taking the modes of $q_\phi(\vz|\rmX)$ and $p_d(\rvx_d; \eta_d(\rvz; \theta))$.

\subsection{Importance weighted autoencoder (IWAE)}

Importance weighted autoencoder (IWAE)~\citep{burda2016importance} differs from VAE only on the training loss.

\paragraph{Loss.} Instead of maximizing the ELBO, IWAE maximizes a tighter loss that makes use of $K$ i.i.d. samples from $\rvz$:
\begin{equation}
    \operatorname{IWAE}(\rmX, \theta, \phi) \coloneqq \Eop[\rvz_1, \dots, \rvz_K \sim q_\phi]{\log \frac{1}{K} \sum_k \frac{p_\theta(\rmX|\rvz_k)p(\rvz_k)}{q_\phi(\rvz_k|\rmX)}}.
\end{equation}
For all the results shown in \cref{tab:test_errors} we set the number of importance samples to $K=20$.

\subsection{Doubly reparametrized gradient estimator (DReG)} \label{subsec:dreg}

\citet{rainforth2018tighter} showed that the gradient estimators produced by IWAE have some undesired properties that could hamper properly learning the inference parameters (encoder).
A strict improvement over this negative result was later provided by \citet{tucker2018dreg}, as they provide a simple way of addressing these issues by applying the reparametrization trick a second time.
As a result, we obtain again a model structurally identical to VAE, but which is optimized with two different losses: one for the encoder, and one for the decoder. 
We use $K=20$ importance samples as for IWAE.

\paragraph{Encoder loss.} For one importance sample $\rvz_k$, let us define 
\begin{equation}
    \omega_k \coloneqq \frac{p_\theta(\rmX|\rvz_k)p(\rvz_k)}{q_\phi(\rvz_k)}\text{, and}\quad\widetilde{\omega}_k \coloneqq \frac{\omega_k}{\sum_i \omega_i}\text{ such that }\quad\sum_k \widetilde{\omega}_k = 1.
\end{equation}
Then, we optimize the parameters of the encoder by maximizing
\begin{equation}
    \operatorname{DReG}^\text{enc}(\rmX, \theta, \phi) \coloneqq \Eop[\rvz_1, \dots, \rvz_K \sim q_\phi]{\sum_k \widetilde{\omega}_k^2 \log \omega_k},
\end{equation}
where we consider $\widetilde{\omega}_k$ to be a constant value (\ie., we do not backpropagate through it), and we compute the derivative \wrt $\phi$ only through $\rvz$ (\ie, we do not compute the partial derivative \wrt $\phi$).

\textbf{Decoder loss.}
Similarly, we optimize the parameters of the decoder by maximizing the following loss (same assumptions on $\widetilde{\omega}_k$ and $\phi$):
\begin{equation}
    \operatorname{DReG}^\text{dec}(\rmX, \theta, \phi) \coloneqq \Eop[\rvz_1, \dots, \rvz_K \sim q_\phi]{\sum_k \widetilde{\omega}_k \log \omega_k}.
\end{equation}

\subsection{HI-VAE}

We have faithfully re-implemented the original version of HI-VAE~\citep{nazabal2020hivae}, this includes implementing their architecture with the same number of parameters, as well as implementing their methods (such as the proposed normalization and denormalization layers).
Regarding the architecture, we have maintained the same one as the original authors used in their experiments. Therefore, results between HI-VAE and the rest of the models in \cref{tab:test_errors} are not completely comparable.

HI-VAE assumes a hierarchical latent space. Thus, we assume the following probabilistic model:
$$
\begin{array}{lrl}
    \text{Prior:} & p(\rvz, \rvs) &= p(\rvs)p(\rvz|\rvs) \\ &&= \text{Cat}(\frac{1}{d_s}, \frac{1}{d_s}, \dots, \frac{1}{d_s})\,\mathcal{N}(\mu_0(\rvs), I) \vspace{1em} \\
    \text{Likelihood:} & p_\theta(\rmX | \rvz) &= \prod_d p_d(\rvx_d | \eta_d(\rvz; \theta)) \vspace{1em} \\ 
    \text{Variational approx.:} & q_\phi(\rvz, \rvs|\rmX) &= q_\phi(\rvs|\rmX)q_\phi(\rvz|\rmX, \rvs) \\ &&= \text{Cat}(\pi(\rmX))\, \mathcal{N}(\mu(\rmX, \rvs; \phi), \sigma(\rmX, \rvs; \phi)).
\end{array}
$$
Similar to VAE, $\mu_0$, $\mu$, and $\sigma$ are all neural networks, and all likelihood parameters $\eta_d$ are jointly modelled by the decoder. 
Note also the introduction of new variables to describe the size of each latent variable, $d_z$ and $d_s$.

We set in our experiments $d_z = d_s = 10$, and the hidden size to $h = 5D$, just as in the original paper.

\paragraph{Loss.} We maximize the ELBO as originally proposed by \citet{nazabal2020hivae}:
\begin{equation}
    \operatorname{ELBO}(\rmX, p_\theta, q_\phi) \coloneqq \Eop[\rvz,\rvs\sim q_\phi]{\log p_\theta(\rmX|\rvz, \rvs)} - \KLop{q_\phi(\rvz, \rvs)}{p(\rvz, \rvs)}.
\end{equation}

\subsection{Mixture-based VAEs}

For the mixture-based models\removed{ (\ie, MVAE, MMVAE, MoPoE)}, we have followed the same architecture and setups as the ones used by \citet{shi2019mmvae,sutter2021mopoe}.
When it comes to different models, we only have changed the way we sample the modalities $\rmZ_A$ by changing the selection of $\mathcal{A}$, but the architectures remain the same as the ones used in previous literature.

\begin{added*}
Therefore, we here describe the architecture for all the models at once, as they differ on the loss function and the experts, which does not modify the underlying network.
We assume the following probabilistic model for the MNIST-SVHN-Text experiments:
\begin{center}
    \def\arraystretch{1.5}%
    \begin{tabular}{lL}
        Prior: & p(\rvz) = \mathcal{N}(0, I)  \\ 
        Likelihood: & p_\theta(\rmX | \rvz) = \text{Laplace}(\rvx_{M} | \mu(\rvz; \theta), 0.75)\;\text{Laplace}(\rvx_{S} | \mu(\rvz; \theta), 0.75)\;\text{Cat}(\rvx_{T} | \pi(\rvz; \theta)) \\ 
        Variational approx.: & q_\phi(\rvz|\rmX) = \mathcal{N}(\mu(\rmX; \phi), \sigma(\rmX; \phi))
    \end{tabular}
\end{center}
where variables are properly transformer to meet their constraints, \eg, we use a softmax to model the class probabilities of the likelihood of the text modality.
We consider the following encoders and decoders for each modality:

\textbf{MNIST:}

\begin{center}
    \begin{tabular}{ll}
        \textbf{Encoder:} & \linear{h}\relu\linear{h}\relu\linear{2\latentsize} \\
        \textbf{Decoder:} & \linear{h}\relu\linear{h}\relu\linear{2D}\Sigmoid \\
    \end{tabular}
\end{center}

\textbf{SVHN:}

\begin{center}
    \begin{tabular}{ll}
        \textbf{Encoder:} & \conv{4}{2}{1}\relu\conv{4}{2}{1}\relu\conv{4}{2}{1}\relu\conv{4}{1}{0}\\
        \textbf{Decoder:} & \convt{4}{1}{0}\relu\convt{4}{2}{1}\relu\convt{4}{2}{1}\relu\conv{4}{2}{1}\Sigmoid\\
    \end{tabular}
\end{center}
where the last convolutional layer of the encoder is repeated twice, one for each parameter of the variational approximation.

\textbf{Text:}

\begin{center}
    \begin{tabular}{ll}
        \textbf{Encoder:} & \conv{1}{1}{0}\relu\conv{4}{2}{1}\relu\conv{4}{2}{0}\relu\linear{2\latentsize} \\
        \textbf{Decoder:} & \linear{D}\convt{4}{1}{0}\relu\convt{4}{2}{1}\relu\conv{1}{1}{0} \\
    \end{tabular}
\end{center}

\textbf{Experimental setup.} For each experiment, we train the model for $30$ epochs and a batch size of $128$. We use AMSGrad~\citep{j.2018on} with a learning rate of $0.001$. Regarding the variational loss, we use $K=30$ importance samples for all losses (when using the ELBO, we instead use those samples for the Monte Carlo estimator of the outer expectation). %
For evaluation, we take the model parameters with the highest validation error (\SI{10}{\percent} of the training data) during training, and report all the metrics with respect to a test set.
\end{added*}
}
\section{Experimental details} \label{app:experimental-setup}
\subsection{Heterogeneous experiments}
\subsubsection{Dataset descriptions}

\textbf{Likelihood selection.} Choosing the proper likelihood is a hard task which requires expert-domain knowledge for each specific setting. 
We attempt to simplify this process, and instead automatize likelihood selection based on basic properties of the data that can be programmatically verified. Specifically, we use the following criteria:
\begin{center}
    \begin{tabular}{lr}
        \textbf{Real-valued:} & $x_d\sim \mathcal{N}(\mu, \sigma)$ \\
        \textbf{Positive real-valued:} & $x_d\sim \log \mathcal{N}(\mu, \sigma)$ \\
        \textbf{Count:} & $x_d\sim \operatorname{Poiss}(\lambda)$ \\ 
        \textbf{Binary:} & $x_d\sim \operatorname{Bern}(p)$ \\
        \textbf{Categorical:} & $x_d\sim \operatorname{Cat}(\pi_1, \pi_2, \dots, \pi_K)$
    \end{tabular}
\end{center}

\paragraph{Datasets.} 
For the experiments shown in \cref{subsec:heterogeneous}, we use 12 different heterogeneous and homogeneous datasets. 
First, we took \textit{Adult}, \textit{defaultCredit}, \textit{Wine}, \textit{Bank marketing},
{\textit{El Nino}, \textit{Magic}, and \textit{MiniBooNE}}
datasets from the UCI repository \citep{Dua:2019}. Then, we included from the R package datasets \citep{Rpackage} the following datasets: \textit{Diamonds}, \textit{Movies (IMDB)}, \textit{Health Insurance (HI)}, \textit{German health registry (rwm5yr)}, and \textit{labour}.
Table \ref{tab:datasets} provides the statistics per dataset in terms of sizes and number of likelihoods. 
It is important to remark that the \textit{IMDB} and \textit{Adult} datasets contain NaNs values (each only in two of the features). We replace them by non-NaNs values and ignore them during training and evaluation using boolean masks (similar to what \citet{nazabal2020hivae} do).

\begin{table}[!hbtp]
    \centering
    \caption{Datasets description. The first two columns describe number of instances, $N$, and number of features, $D$. The next columns describe the number of data types per dataset. {Note that the last three datasets are homogeneous, and thus only have real variables}.
    } \label{tab:datasets}
    \begin{tabular}{l|cc|cccc} \toprule
        \textit{Dataset} & $N$ & $D$ & \textit{Real} & \textit{Positive} & \textit{Count} & \textit{Categorical} \\
        \midrule
        Adult & 32561 & 12 & 0 & 3 & 1 & 7 \\
        Credit & 30000 & 24 & 6 & 7 & 1 & 10 \\
        Wine & 6497 & 13 & 0 & 11 & 1 & 1 \\
        Diamonds & 53940 & 10 & 7 & 0 & 0 & 3 \\
        Bank & 41188 & 21 & 10 & 0 & 0 & 11 \\
        IMDB & 28819 & 23 & 4 & 1 & 10 & 8 \\
        HI & 22272 & 12 & 5 & 1 & 0 & 6 \\
        rwm5yr & 19609 & 16 & 0 & 2 & 3 & 11 \\
        labour & 15992 & 9 & 3 & 0 & 2 & 4 \\
        \midrule
        El Nino & 178080 & 12 & 12 & 0 & 0 & 0 \\
        Magic & 19020 & 11 & 11 & 0 & 0 & 0 \\
        BooNE & 130065 & 43 & 43 & 0 & 0 & 0 \\
        \bottomrule
    \end{tabular}
\end{table}

\paragraph{Preprocessing.} 
When parsing the dataset, we center all real-valued features by removing their mean.
We further standardize real-valued features, computing their (training) standard deviation and dividing the data by this quantity. 
We also divide by the standard deviation for positive real-valued features (but in the log-space, as we assume a log-normal likelihood).
These last two steps are omitted for HI-VAE, since it uses its own normalization layer as described by~\citet{nazabal2020hivae}.
We also treat non-negative as positive real-valued features by adding a negligible value of \num{1e-20}.
Finally, we make sure that the support of count, binary, and categorical features are in accordance to that of the library used during implementation by removing their minimum value in the case of binary and categorical features, and $1$ in the case of count features.

Additionally, we performed some extra preprocessing to the \textit{IMDB} {and \textit{Bank}} datasets. 
{In the \textit{IMDB} dataset, }%
there are ten features that contain rating percentages of users to the movies, ranging from \num{0} to \num{100}, at intervals of $0.5$. We convert each of them into discrete features starting from one by performing $\rvx_d' = 2\rvx_d+1$ to each of these features, treating them afterwards as count data. 
{As for the \textit{Bank} dataset, we remove the uninformative dimension 12-th as a data cleaning step.}

\subsubsection{Experimental settings}

We train all experiments using Adam as optimizer, with a learning rate of \num{0.001} for all models. 
For all models (except HI-VAE) we set the batch size to \num{128},
and train for \num{400} epochs for the all datasets (except for \textit{Wine} with \num{2000} epochs).
For HI-VAE, we set the batch size to \num{1000} and the number of epochs to \num{2000} as in the original paper. 
We randomly split the data into training (\SI{70}{\percent}), validation~(\SI{10}{\percent}), and testing (\SI{20}{\percent}).
We set the latent size of $\rvz$, $d$, to \SI{50}{\percent} of the number of features of the dataset, $D$, and the hidden size of each layer to \num{50} for all the experiments, except for those of the \textit{Bank} dataset which are set to \num{100}.

\textbf{Metric.} 
Since we deal with heterogeneous data, where each feature has different type and range, we compute the reconstruction error using metrics that account for these differences. 
For numerical features (real, positive, and count data) we compute the normalized root mean squared error:
\begin{equation}
    \operatorname{err}(d) = \frac{1}{N}\frac{\norm[2]{x_d - \hat{x}_d}}{\max{(x_d)} - \min{(x_d)}},
\end{equation}
where $\hat{x}$ is the model prediction. 
For the case of nominal features (categorical and binary data) we use the error rate as reconstruction error:
\begin{equation}
    \operatorname{err}(d) = \frac{1}{N} \sum_{n=1}^N I(x_{n,d} \neq \hat{x}_{n,d}).
\end{equation}
The final metric shown in \cref{tab:test_errors} is the average across dimensions, $\operatorname{err} = \frac{1}{D} \sum_d \operatorname{err}(d)$.

\paragraph{Model selection.}
In order to make fair comparisons, for each model and dataset we first tuned the hyperparameters (for example, hidden/latent/batch size, number of epochs, etc.)
for the vanilla implementations (i.e., without 
modifying the backward pass).
To this end, we ran grid searches and averaged the validation metric over five random seeds, just as in \cref{tab:test_errors}, choosing the set of hyperparameters that performed the best in terms of reconstruction error during validation.
Note that all these hyperparameters (including optimization hyperparameters such as learning rate) are shared across all methods of the same setting.
Additionally, we verified that the vanilla models were performing well by visually inspecting the marginal reconstructions.

\paragraph{{Selecting the algorithm $f_\psi$.}}
For the heterogeneous experiments we trained all the possible combinations between the following magnitude-aware algorithms: \{nothing,~GradNorm~\citep{chen2017gradnorm}, ~MGDA-UB~\citep{mgdaub}, ~\hbox{IMTL-G}~\citep{liu2021imtl}\}
and direction-aware algorithms:\{nothing, ~GradDrop~\citep{graddrop}, ~PCGrad~\citep{yu2020pcgrad}\} on the training data. This amounts to a total of \num{12} combinations, plus the hyperparameter of specific algorithms. In this case, we only tune the $\alpha$ parameter from GradNorm between the values zero and one. Then, similar to model selection, we chose the best {algorithm} %
 by averaging over five random seeds and taking the combination of methods that performed the best in terms of reconstruction error in validation {(see \cref{tab:bestMTL})}.
In general, it was enough to focus on the median to select the best combination. However, some combinations had outliers, and we chose those having a good balance between median, mean, and standard deviation.

\sisetup{table-format=1.2, round-mode = places, round-precision = 2, detect-all}
\begin{table*}[t]
    \setlength\tabcolsep{2.4pt}
    \centering
    \caption{The best MTL methods chosen by cross-validation. GN, GD, PG, and MGDA stand for GradNorm, GradDrop, PCGrad, and MGDA-UB, respectively.} \label{tab:bestMTL}
    {
        \begin{tabular}{l|llll}
\toprule
\textit{Dataset} & VAE-ELBO & VAE-IWAE & VAE-DReG & HI-VAE \\
\midrule
Adult & IMTL-G & IMTL-G &        IMTL-G-PG & GN-PG ($\alpha=1$) \\
Credit &                 IMTL-G &        IMTL-G-GD &      IMTL-G-GD &  GN ($\alpha=1$) \\
Wine &          GN ($\alpha=0$) &   GN-PG ($\alpha=0$) &  GN ($\alpha=0$) &        GN ($\alpha=0$) \\
Diamonds & IMTL-G &  IMTL-G &  IMTL-G-PG &        GN ($\alpha=0$) \\
Bank &  GN ($\alpha=0$) & GN-GD ($\alpha=0$) & GN ($\alpha=0$) &  MGDA-PG \\
IMDB & GN-GD ($\alpha=0$) &  GN ($\alpha=0$) & GN-PG ($\alpha=0$) & GN-PG ($\alpha=0$) \\
HI & GN-GD ($\alpha=0$) & GN ($\alpha=0$) & GN-PG ($\alpha=0$) &  MGDA \\
rwm5yr &  GN ($\alpha=1$) & GN-GD ($\alpha=1$) &  GN ($\alpha=1$) & MGDA-PG \\
labour & GN ($\alpha=0$) & GN ($\alpha=1$) & GN-PG ($\alpha=0$) & GN ($\alpha=0$) \\
El Nino &  IMTL-G &  IMTL-G-PG &  IMTL-G-GD &        GN ($\alpha=0$) \\
Magic &  GN ($\alpha=1$) & IMTL-G &        GN ($\alpha=1$) &  IMTL-G \\
BooNE &  IMTL-G-PG &   IMTL-G & GN-PG ($\alpha=0$) &   MGDA-PG \\
\bottomrule
\end{tabular}
    }
\end{table*}

\paragraph{Statistical test.}
In order to compare the performance of the proposed method with the baseline, we employ the corrected paired t-test~\citep{corrected-t-test}. 
The usual paired t-test assumes that the data used to perform the test is independently sampled, which usually does not hold in the machine learning as we sample the training and test data from the same distribution. 
As a consequence, paired t-test might suggest statistical significance between the compared models, whereas there is no such significance (type I error).
Corrected paired t-test considers the dependency of the sampled data, correcting the variance of the differences of the paired samples in the two testing models.

{
\paragraph{Data Generation.}
To generate the data for the experiments in \cref{subsec:heterogeneous}, we followed the same approach as \citet{ghosh2019variational} and made use of post-hoc Gaussian Mixture Models (GMMs) to approximate the aggregated posterior, $\Eop[\rmX]{q_\phi(\rmZ|\rmX)}$.  
After training the VAE models, we use the latent space $Z$ generated from the training data and fit a GMM (with 100 components) on that data. Next, we use this GMM to sample a dataset with as many samples as the test data. %

}
\subsubsection{Additional experimental results}

\begin{figure}[t]
    \centering
    \includegraphics[width=\linewidth, keepaspectratio]{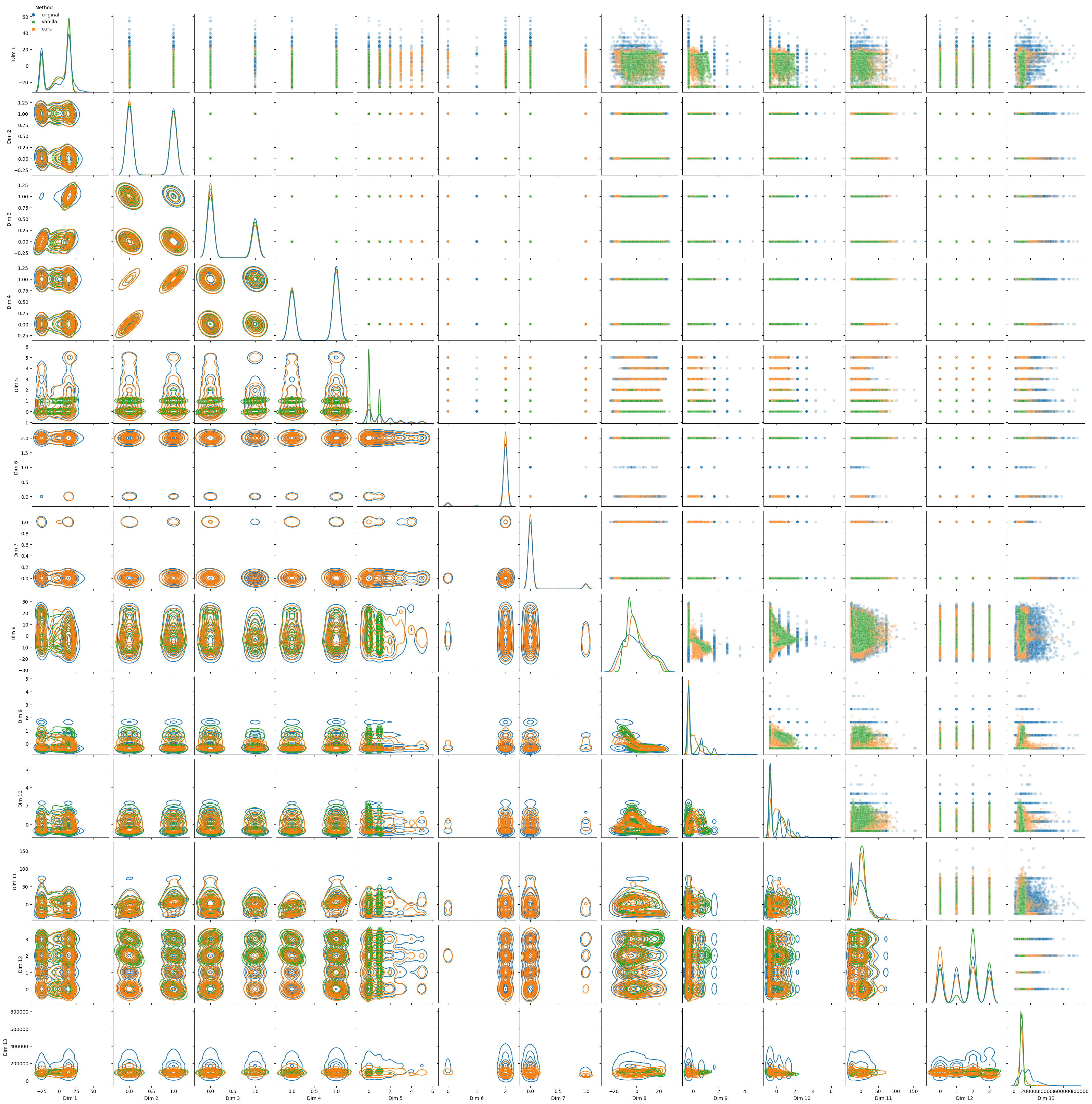}
    \caption{Pair plot of all the dimensions of \textit{HI}, generated from different VAE models. Diagonal show the marginals, upper-diagonals scatter plots, and lower-diagonals kernel density estimates. The VAE trained with our approach is able to generate faithful samples.}
    \label{fig:pairplot1}
\end{figure}

\begin{figure}[t]
    \centering
    \includegraphics[width=\linewidth, keepaspectratio]{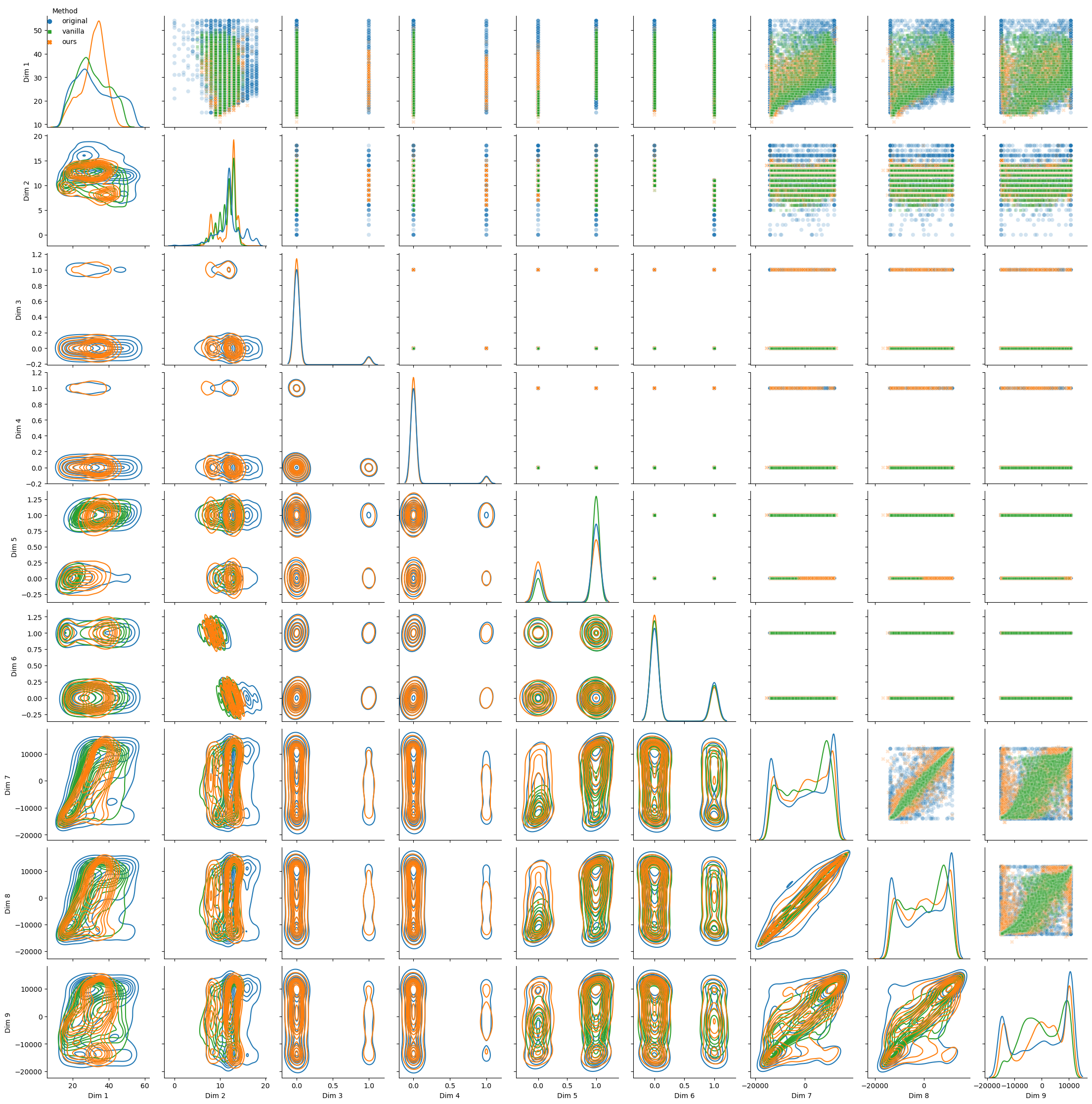}
    \caption{Pair plot of all the dimensions of \textit{labour}, generated from different VAE models. Diagonal show the marginals, upper-diagonals scatter plots, and lower-diagonals kernel density estimates. The VAE trained with our approach is able to generate faithful samples.}
    \label{fig:pairplot2}
\end{figure}

In addition to the results presented in the main paper, we present in \cref{tab:test_errors_meanstd} the same table as \cref{tab:test_errors} but showing also the standard deviation of the results.
Moreover, we show in \cref{fig:pairplot1,fig:pairplot2} the full pair plot for the \textit{HI} dataset, as well as another full pair plot of the \textit{labour} dataset.

\begin{table}[!hbtp]
    \centering
    \caption{Test reconstruction errors (mean and standard deviation) of different models and losses for the baseline and our framework.} \label{tab:test_errors_meanstd}
   \begin{tabular}{cl|cccc}
\toprule
\textit{Dataset} & \textit{Method} & {VAE-ELBO} & {VAE-IWAE} & {VAE-DReG} & {HI-VAE} \\
\midrule
        \multirow{2}{*}{{\textit{Adult}}} & vanilla & 0.21 $\pm$  0.01 & 0.22 $\pm$  0.02 & 0.24 $\pm$  0.01 & 0.13 $\pm$  0.00 \\
         &    \ours & \textbf{0.11} $\pm$ 0.02 & \textbf{0.12} $\pm$ 0.02 & 0.19 $\pm$ 0.08 & \textbf{0.09} $\pm$ 0.02 \\
         \midrule
\multirow{2}{*}{{\textit{defaultCredit}}} & vanilla & 0.13 $\pm$ 0.00 & 0.14 $\pm$ 0.02 & 0.14 $\pm$ 0.01 & 0.15 $\pm$ 0.09 \\
 &    \ours & \textbf{0.04} $\pm$ 0.00 & \textbf{0.05} $\pm$ 0.01 & \textbf{0.08} $\pm$ 0.01 & 0.06 $\pm$ 0.01 \\
 \midrule
\multirow{2}{*}{{\textit{Wine}}} & vanilla & 0.09 $\pm$ 0.00 & 0.08 $\pm$ 0.00 & 0.08 $\pm$ 0.00 & 0.13 $\pm$ 0.01 \\
 &    \ours & \textbf{0.07} $\pm$ 0.01 & \textbf{0.07} $\pm$ 0.00 & \textbf{0.07} $\pm$ 0.00 & \textbf{0.11} $\pm$ 0.02 \\
 \midrule
\multirow{2}{*}{{\textit{Diamonds}}} & vanilla & 0.19 $\pm$ 0.01 & 0.18 $\pm$ 0.01 & 0.18 $\pm$ 0.00 & 0.11 $\pm$ 0.02 \\
      &    \ours & \textbf{0.13} $\pm$ 0.02 & \textbf{0.12} $\pm$ 0.01 & \textbf{0.14} $\pm$ 0.01 & \textbf{0.01} $\pm$ 0.01 \\
      \midrule
\multirow{2}{*}{{\textit{Bank}}} & vanilla & 0.20 $\pm$ 0.00 & 0.20 $\pm$ 0.00 & 0.19 $\pm$ 0.00 & 0.13 $\pm$ 0.02 \\
          &    \ours & \textbf{0.04} $\pm$ 0.00 & 0.10 $\pm$ 0.05 & 0.11 $\pm$ 0.04 & \textbf{0.10} $\pm$ 0.01 \\
\midrule
\multirow{2}{*}{{\textit{IMDB}}} & vanilla & 0.09 $\pm$ 0.02 & 0.10 $\pm$ 0.02 & 0.10 $\pm$ 0.02 & 0.08 $\pm$ 0.00 \\
&    \ours & \textbf{0.05} $\pm$ 0.04 & \textbf{0.05} $\pm$ 0.04 & \textbf{0.06} $\pm$ 0.04 & 0.10 $\pm$ 0.09 \\
\midrule
\multirow{2}{*}{{\textit{HI}}} & vanilla & 0.17 $\pm$ 0.01 & 0.16 $\pm$ 0.00 & 0.15 $\pm$ 0.00 & 0.11 $\pm$ 0.00 \\
&    \ours & \textbf{0.04} $\pm$ 0.00 & \textbf{0.04} $\pm$ 0.00 & \textbf{0.04} $\pm$ 0.00 & \textbf{0.11} $\pm$ 0.01 \\
\midrule
\multirow{2}{*}{{\textit{rwm5yr}}} & vanilla & 0.11 $\pm$ 0.01 & 0.09 $\pm$ 0.00 & 0.10 $\pm$ 0.00 & 0.04 $\pm$ 0.01 \\
&    \ours & \textbf{0.03} $\pm$ 0.00 & \textbf{0.03} $\pm$ 0.01 & \textbf{0.03} $\pm$ 0.00 & \textbf{0.02} $\pm$ 0.00 \\
\midrule
\multirow{2}{*}{{\textit{labour}}} & vanilla & 0.11 $\pm$ 0.00 & 0.10 $\pm$ 0.00 & 0.10 $\pm$ 0.00 & 0.10 $\pm$ 0.00 \\
&    \ours & \textbf{0.06} $\pm$ 0.00 & \textbf{0.07} $\pm$ 0.00 & \textbf{0.08} $\pm$ 0.01 & \textbf{0.07} $\pm$ 0.00 \\
\midrule \midrule
\multirow{2}{*}{{\textit{EL Nino}}} & vanilla & 0.10 $\pm$ 0.01 & 0.09 $\pm$ 0.00 & 0.08 $\pm$ 0.00 & 0.10 $\pm$ 0.01 \\
&    \ours & \textbf{0.07} $\pm$ 0.01 & \textbf{0.06} $\pm$ 0.01 & \textbf{0.07} $\pm$ 0.00 & \textbf{0.02} $\pm$ 0.00 \\
\midrule
\multirow{2}{*}{{\textit{Magic}}} & vanilla & 0.06 $\pm$ 0.00 & 0.05 $\pm$ 0.00 & 0.05 $\pm$ 0.00 & 0.06 $\pm$ 0.00 \\
         &    \ours & 0.06 $\pm$ 0.00 & 0.05 $\pm$ 0.00 & 0.05 $\pm$ 0.00 & \textbf{0.03} $\pm$ 0.00 \\
\midrule
\multirow{2}{*}{{\textit{BooNE}}} & vanilla & 0.04 $\pm$ 0.00 & 0.04 $\pm$ 0.00 & 0.04 $\pm$ 0.00 & 0.04 $\pm$ 0.00 \\
&    \ours & 0.04 $\pm$ 0.00 & \textbf{0.04} $\pm$ 0.00 & 0.04 $\pm$ 0.00 & 0.04 $\pm$ 0.00 \\
\bottomrule
\end{tabular}
\end{table}

\subsection{Multimodal experiments}

\subsubsection{Experiment details}

For the multimodal experiments on MNIST-SVHN-Text, we have followed the same setup (including hyper-parameters) as \citet{shi2019mmvae} and \citet{sutter2021mopoe}.
We differ from their setups in that, in order to provide a fair comparison between losses, we always employ $K=30$ samples from $\rmZ$, whether they are used as importance samples (IWAE, SIWAE) or used for the Monte-Carlo approximation of the expected value \wrt $\rmZ$.
Also, we do model selection using a validation dataset ($10\%$ of the training data), and use a test set to obtain all the results presented in this work.
Following \citet{shi2019mmvae}, we use the Sticking-The-Landing estimator (STL) \citep{Roeder2017StickingTL} for all losses. In short, this estimator simply omits the partial derivatives of the variational approximation \wrt the encoder parameters. Note that \citet{shi2019mmvae} did not mention this estimator, but they rather talk about the DReG loss~\citep{tucker2018dreg}. However, due to a bug in their code, they effectively compute the STL estimator in their experiments.

\paragraph{Selecting algorithm $f_\psi$. }
Since the number of \block{s} is large, and the training times are considerably longer than for the heterogeneous experiments, here we keep performing cross-validation, but this time we substitute grid-selection by hand-picked hyperparameters options that we observed to perform better than others (for example, we replaced IMTL-G~\cite{liu2021imtl} by CAGrad~\cite{cagrad}, as it was really clear by looking at the logs that IMTL-G was not working at all).
Instead of looking for a specific algorithm for each of the \block{s}, we assume the same algorithm for all of them (same hyperparameters, but different parameters) and only cross-validate by using the modified backward pass on the blocks associated with the different goals in an incremental way (\ie, as presented in \replaced{\cref{tab:times} in}{the inset table of \cref{sec:experiments} in} the main paper).

\added{
Choosing the best algorithm in the multimodal setup is more complicated, as we care about different metrics (coherence and latent classification) at different levels (self and cross metrics), for each modality.
We group all metrics in metric-type pairs (\eg, latent-classification-self), and within each group, we group them by the expert/modality they are testing (\eg, cross latent classification for the first expert tests all other latent samples in the classifier of the first expert).
For each metric, we compute a value $I(x, y)$, where $x$ is the value obtained by the algorithm, and $y$ the value obtained by the baseline, and take the average of each sequence recursively until obtaining a single number.
We use the relative improvement $I(x, y) = \Delta(x,y) = \frac{x - y}{y}$ to compare the different metrics, choosing the method that obtains the best improvement, averaged across experts/modalities and metrics.
For MVAE, we noticed that the metrics tend to oscillate and there are important trade-offs in performance. Therefore, for this model we adopt a more conservative approach and use $I(x,y) = \mathbf{1}_{x \geq y}$, to choose the algorithm that, on average, improves the most number of metrics.
}

\subsection{Additional experimental results}

\begin{figure*}
	\centering
    \hfill %
    \begin{subfigure}[c]{.33\textwidth}
        \includegraphics[keepaspectratio, width=\linewidth]{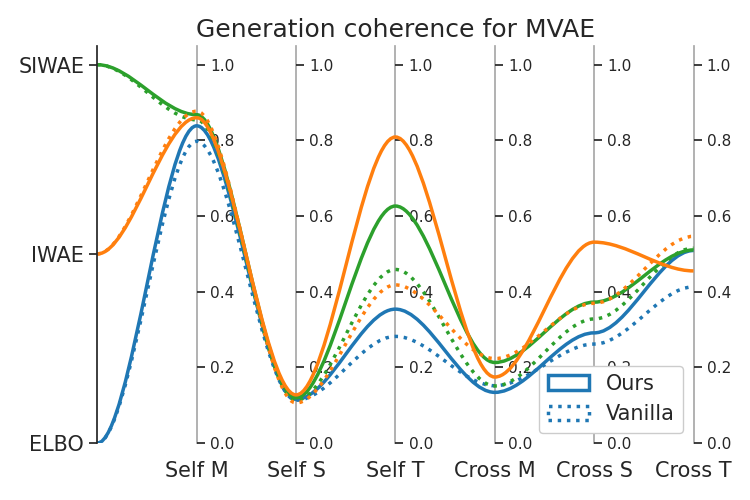} 
        \caption{MVAE.}
    \end{subfigure} %
    \hfill %
    \begin{subfigure}[c]{.33\textwidth}
        \includegraphics[keepaspectratio, width=\linewidth]{figs/mmvae_parallel.png} 
        \caption{MMVAE.}
    \end{subfigure} %
    \hfill
    \begin{subfigure}[c]{.33\textwidth}
        \includegraphics[keepaspectratio, width=\linewidth]{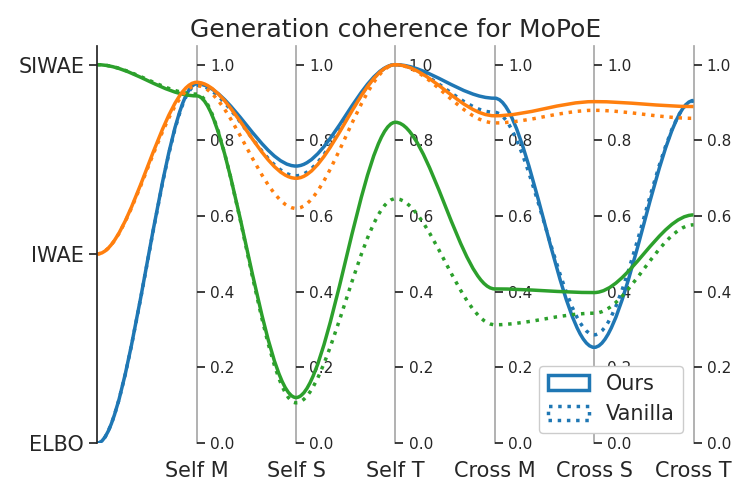} 
        \caption{MoPoE.}
    \end{subfigure} %
    \caption{Generation coherence results for all models and losses. In general, improve all metrics \wrt the baseline.}
    \label{fig:all-parallel-full}
\end{figure*}

In this section, we have included the complete results for the MNIST-SVHN-Text experiments. Specifically, we present: the reconstruction coherence results for all the three losses (\cref{tab:multimodal-reconstruction-full}); the self and cross coherence results in tabular form for the three losses (\cref{tab:multimodal-elbo-full,tab:multimodal-iwae-full,tab:multimodal-siwae-full}), including extra information like the training times, the specific MTL algorithms used, and the goals for which we apply them; 
\added{and the log-likelihoods conditioned on different modalities (\cref{tab:log-lik-elbo,tab:log-lik-iwae,tab:log-lik-siwae}), showing standard deviations as space permits, thus showing the high variance that the vanilla approach shows at times (for example, MoPoE in \cref{tab:log-lik-siwae}).}
Finally, we present the parallel coordinate plots for the three models (\cref{fig:all-parallel-full}).

{

\sisetup{table-format=2.2, round-mode = places, round-precision = 2, detect-all}
\begin{table}
    \centering
    \caption{Reconstruction coherence ($A = \{M, S, T\}$) for each modality, model, and dataset.}
    \label{tab:multimodal-reconstruction-full}
    {
        \begin{tabular}{cl|SSS|SSS|SSS} 
            \toprule 
            & & \multicolumn{3}{c|}{ELBO} & \multicolumn{3}{c|}{IWAE} & \multicolumn{3}{c}{SIWAE} \\ \midrule
            & $\rvx_d$ & {M} & {S} & {T} & {M} & {S} & {T} & {M} & {S} & {T} \\ \midrule
            \multirow{2}{*}{MVAE} & vanilla & 97.5285804271698 & 88.26178431510925 & 99.30322647094727 & 97.26999998092651 & 87.19438552856446 & 98.75560164451599 & 97.37306952476501 & 87.47473120689392 & 98.82936596870422 \\
            & \ours & \bfseries 97.85028994083405 & \bfseries 89.65040445327759 & \bfseries 99.63676035404205 & \bfseries 98.27756136655807 & \bfseries 89.0097588300705 & \bfseries 99.92557913064957 & 97.41689711809158 & 87.62706816196442 & \bfseries 99.19884651899338 \\ \midrule
            \multirow{2}{*}{MMVAE} & vanilla & 86.01003438234329 & 45.58732658624649 & 89.16780799627304 & 85.24692058563232 & 84.03324335813522 & 88.66259902715683 & 58.94593670964241 & 61.26817092299461 & 63.27350363135338 \\
            & \ours & \bfseries 89.4158273935318 & 45.83193063735962 & \bfseries 91.53931885957718 & 87.55284349123636 & \bfseries 86.87140742937723 & 90.93259473641714 & \bfseries 74.84600245952606 & \bfseries 73.89406164487204 & \bfseries 81.09348515669504 \\ \midrule
            \multirow{2}{*}{MoPoE} & vanilla & 95.71546614170074 & 85.86411327123642 & 98.0124294757843 & 95.82313895225525 & 87.55003064870834 & 97.9305237531662 & 75.09585581719875 & 67.16080456972122 & 76.60753987729549 \\
            & \ours & \bfseries 96.50165736675262 & \bfseries 93.60147714614868 & \bfseries 99.14027452468872 & \bfseries 97.29161858558655 & \bfseries 92.93432831764221 & \bfseries 98.99884462356567 & 96.91308736801147 & 89.01377469301224 & 99.28467273712158 \\
            \bottomrule
        \end{tabular}
    }
\end{table}

}

{

\sisetup{table-format=2.2, round-mode = places, round-precision = 2, detect-all}
\begin{table*}
    \centering
    \caption{Self and cross generation coherence (\%) results for different models on \textbf{M}NIST-\textbf{S}VHN-\textbf{T}ext, trained using ELBO and averaged over 5 different seeds. Models trained with our framework are able to sample more coherent modalities.} \label{tab:multimodal-elbo-full}
    \resizebox{\textwidth}{!}
    {
        \begin{tabular}{cl|ccc|SSS[table-format=3.2]|SSS|SSS|SSS|S} 
            \toprule 
            & & & & & \multicolumn{3}{c}{Self coherence} & \multicolumn{9}{c|}{Cross coherence} & \\
            & $\vx_d$ & & & & {M} & {S} & {T} & \multicolumn{3}{c}{M} & \multicolumn{3}{c}{S} & \multicolumn{3}{c|}{T} & \text{Time}  \\
            & $A$ & \ac*{LI} & \ac*{EEI} & \ac*{DEI} & {M} & {S} & {T} & {S} & {T} & \multicolumn{1}{c|}{S,T} & {M} & {T} & \multicolumn{1}{c|}{M,T} & {M} & {S} & \multicolumn{1}{c|}{M,S} & h \\ \midrule
            \multirow{2}{*}{MVAE}  & {vanilla} & \openbox & \openbox & \openbox & 80.29846668243408 & 12.6258210837841 & 25.77445864677429 & 11.121198534965511 & 18.21829058229923 & 19.4893579185009 & 43.293524980545045 & 18.497849553823464 & 16.900739222764966 & 51.82091534137726 & 11.651328951120372 & 54.71137940883637 & 3.8223888888888893 \\
            & {CG~($\alpha=0.4$)} & \checkbox & \openbox & \openbox & 85.12432426214218 & 12.342862784862515 & 34.637197107076645 & 10.69862395524978 & 12.51399256289005 & 16.82608518749475 & 44.94035840034485 & 22.359844297170635 & \bfseries 29.451550543308258 & 61.41359210014343 & 12.256772257387635 & 69.7690412402153 & 4.8229861111111114 \\\midrule
            \multirow{2}{*}{MMVAE} & {vanilla} & \openbox & \openbox & \openbox & \bfseries 95.2276736497879 & 68.2533785700798 & 99.98852461576462 & 62.9275381565094 & 99.9227300286293 & 81.42689913511276 & 31.05553761124611 & 37.42483928799629 & 34.23593267798424 & 96.27213478088379 & 71.28673940896988 & 83.78856480121613 & 10.069791666666667 \\
            & {CA~($\alpha=10.0$)} & \checkbox & \checkbox & \checkbox & 92.62228161096573 & \bfseries 73.83767068386078 & 99.98554736375809 & \bfseries 76.00729763507843 & 99.59245920181274 & \bfseries 87.79663294553757 & 29.11144271492958 & 34.45761352777481 & 31.780608743429184 & 95.26804834604263 & \bfseries 79.34017926454544 & \bfseries 87.30775266885757 & 13.05798611111111 \\ \midrule
            \multirow{2}{*}{MoPoE} & {vanilla} & \openbox & \openbox & \openbox & 94.52371001243591 & 71.20701223611832 & 99.98939352273941 & 66.80333912372589 & 99.9794065952301 & 98.12058955430984 & \bfseries 19.697178527712822 & 33.247241377830505 & 31.55098557472229 & \bfseries 96.79142832756042 & 77.04015821218491 & \bfseries 97.15262800455093 & 23.244930555555555 \\
            & {CG~($\alpha=10$)} & \checkbox & \checkbox & \checkbox & 94.64496523141861 & \bfseries 73.35900366306305 & \bfseries 100.0 & \bfseries 74.75752681493759 & 99.97854679822922 & 98.42367023229599 & 15.8425323665142 & 32.836320251226425 & 31.03411868214607 & 96.03300988674164 & \bfseries 78.60849350690842 & 96.66668474674225 & 29.412777777777777 \\
            \bottomrule
        \end{tabular}
    }
\end{table*}

}
{

\sisetup{table-format=2.2, round-mode = places, round-precision = 2, detect-all}
\begin{table*}
    \centering
    \caption{Self and cross generation coherence (\%) results for different models on \textbf{M}NIST-\textbf{S}VHN-\textbf{T}ext, trained using IWAE and averaged over 5 different seeds. Models trained with our framework are able to sample more coherent modalities.} \label{tab:multimodal-iwae-full}
    \resizebox{\textwidth}{!}
    {
        \begin{tabular}{cl|ccc|SSS|SSS|SSS|SSS|S} 
            \toprule 
            & & & & & \multicolumn{3}{c}{Self coherence} & \multicolumn{9}{c|}{Cross coherence} & \\
            & $\vx_d$ & & & & {M} & {S} & {T} & \multicolumn{3}{c}{M} & \multicolumn{3}{c}{S} & \multicolumn{3}{c|}{T} & \text{Time}  \\
            & $A$ & \ac*{LI} & \ac*{EEI} & \ac*{DEI} & {M} & {S} & {T} & {S} & {T} & \multicolumn{1}{c|}{S,T} & {M} & {T} & \multicolumn{1}{c|}{M,T} & {M} & {S} & \multicolumn{1}{c|}{M,S} & h \\ \midrule
            \multirow{2}{*}{MVAE} & {vanilla} & \openbox & \openbox & \openbox & 87.32664823532105 & 11.699514538049693 & 37.49252915382385 & 10.758992880582804 & \bfseries 26.63348317146301 & 29.478434324264524 & 54.09736335277557 & 25.163763910532 & 29.506842643022535 & \bfseries 70.24494886398316 & 11.379081755876538 & 72.32779860496521 & 3.7995555555555556 \\
            & {GN~($\alpha=0.0$)} & \checkbox & \openbox & \openbox & 85.72510629892349 & 12.67670691013336 & \bfseries 79.19154316186905 & 11.178777925670143 & 19.452513381838796 & 22.049068659543986 & 49.3730753660202 & \bfseries 55.8295913040638 & \bfseries 54.47539612650871 & 59.55990999937057 & 11.791023612022397 & 63.68505656719208 & 4.236666666666667 \\ \midrule
            \multirow{2}{*}{MMVAE}& {vanilla} & \openbox & \openbox & \openbox & 94.69747692346573 & 68.13080459833145 & 99.9903604388237 & 62.34240159392357 & 98.79158288240433 & 80.55326640605927 & 86.71853095293045 & 97.2883328795433 & 92.00959354639053 & \bfseries 96.82694971561432 & 69.2291796207428 & 83.02458375692368 & 10.075486111111111 \\
            & {CG~($\alpha=0.4$)} & \checkbox & \openbox & \openbox & 94.96151606241861 & 73.66157273451486 & 99.99153117338815 & 68.48965386549631 & \bfseries 99.24991031487784 & 83.84911815325418 & \bfseries 88.98889621098837 & \bfseries 97.97220627466837 & \bfseries 93.48858892917633 & 96.26948237419128 & 76.54984692732492 & 86.40090028444925 & 11.95814814814815 \\\midrule
            \multirow{2}{*}{MoPoE} & {vanilla} & \openbox & \openbox & \openbox & 94.44483518600464 & 63.16262483596802 & 99.97405260801316 & 58.200603723526 & 99.009869992733 & 99.24275726079941 & 80.75450658798218 & 94.76495534181595 & 88.62375020980835 & 96.51471227407455 & 66.13101065158844 & 96.03478312492371 & 23.2675 \\
            & {CG~($\alpha=10.0$)} & \checkbox & \checkbox & \checkbox & \bfseries 95.40619552135468 & \bfseries 69.54438239336014 & \bfseries 99.98906420707703 & 61.749838292598724 & 99.064502120018 & 99.06584024429321 & 81.22643232345581 & \bfseries 96.4717224240303 & \bfseries 92.98069477081299 & 96.6359555721283 & \bfseries 73.32838028669357 & 96.16186022758484 & 29.347152777777774 \\
            \bottomrule
        \end{tabular}
    }
\end{table*}

}
{

\sisetup{table-format=2.2, round-mode = places, round-precision = 2, detect-all}
\begin{table*}
    \centering
    \caption{Self and cross generation coherence (\%) results for different models on \textbf{M}NIST-\textbf{S}VHN-\textbf{T}ext, trained using SIWAE and averaged over 5 different seeds. Models trained with our framework are able to sample more coherent modalities.} \label{tab:multimodal-siwae-full}
    \resizebox{\textwidth}{!}
    {
        \begin{tabular}{cl|ccc|SSS|SSS|SSS|SSS|S} 
            \toprule 
            & & & & & \multicolumn{3}{c}{Self coherence} & \multicolumn{9}{c|}{Cross coherence} & \\
            & $\vx_d$ & & & & {M} & {S} & {T} & \multicolumn{3}{c}{M} & \multicolumn{3}{c}{S} & \multicolumn{3}{c|}{T} & \text{Time}  \\
            & $A$ & \ac*{LI} & \ac*{EEI} & \ac*{DEI} & {M} & {S} & {T} & {S} & {T} & \multicolumn{1}{c|}{S,T} & {M} & {T} & \multicolumn{1}{c|}{M,T} & {M} & {S} & \multicolumn{1}{c|}{M,S} & h \\ \midrule
            \multirow{2}{*}{MVAE}& {vanilla} & \openbox & \openbox & \openbox & 82.06321358680725 & 12.078495621681211 & 36.668736934661866 & 10.342429280281062 & 17.118748845532533 & 19.186700284481048 & 49.992209672927856 & 19.312240332365033 & 31.188021302223207 & 62.50251233577728 & 10.815181583166118 & 64.25450205802917 & 3.809166666666667 \\
            & {CG~($\alpha=0.4$)} & \checkbox & \openbox & \openbox & 86.88859045505524 & 12.89413925260305 & \bfseries 59.14595350623131 & 10.99016740918159 & 20.91808170080185 & \bfseries 30.335458368062973 & 54.82720881700516 & \bfseries 31.816108897328377 & 24.968290328979492 & 71.86760306358337 & 11.104153469204898 & 72.74188995361328 & 4.770208333333334 \\ \midrule
            \multirow{2}{*}{MMVAE} & {vanilla} & \openbox & \openbox & \openbox & \bfseries 95.89911252260208 & 48.302508518099785 & 53.02141793072224 & 28.43314968049526 & 52.51514203846455 & 40.45413397252559 & 84.44346040487289 & 51.07617117464541 & 67.76809990406036 & \bfseries 96.80366516113281 & 39.96035195887089 & 68.38035881519318 & 10.067083333333333 \\
            & {CG~($\alpha=0.4$)} & \checkbox & \openbox & \openbox & 95.89718679587045 & 58.19725841283798 & \bfseries 88.69847655296326 & \bfseries 49.32506904006004 & \bfseries 79.31589980920157 & \bfseries 64.30353770653406 & \bfseries 87.28511532147726 & \bfseries 76.17471218109131 & \bfseries 81.7105770111084 & 96.69901430606842 & 57.86134203275045 & 77.27593878904977 & 12.297916666666667 \\ \midrule
            \multirow{2}{*}{MoPoE} & {vanilla} & \openbox & \openbox & \openbox & 92.31970608234406 & 11.597061343491073 & 69.05172914266586 & 10.13441793620586 & 51.02324187755585 & 34.667880460619926 & 41.930220276117325 & 46.385733783245087 & 51.58409625291824 & 85.18766462802887 & 10.566544160246848 & 67.53952503204346 & 23.29472222222222 \\
            & {GN~($\alpha=0.5$)} & \checkbox & \openbox & \openbox & 90.99198579788208 & 11.996489390730854 & \bfseries 83.81824642419815 & \bfseries 10.6312271207571 & 62.75309696793556 & 52.08384543657303 & 28.190246224403376 & 46.90524935722351 & 43.34438294172287 & 79.64047342538834 & 10.81472355872392 & 90.32563269138336 & 24.57597222222222 \\
            \bottomrule
        \end{tabular}
    }
\end{table*}

}

{
\sisetup{table-format=+2.2, round-mode = places, round-precision = 2, detect-all, table-align-uncertainty=true,
            separate-uncertainty=true}

\begin{table}[h]
    \centering
     \caption{Log-likelihood of the joint generative model, conditioned on the variational posterior of subsets of the modalities. 
     Results report on ELBO as loss function and are averaged over 5 different seeds. Each log-likelihood is divided by the dimensionality of its modality before adding them up, to better reflect improvement across modalities.} \label{tab:log-lik-elbo}
    \resizebox{\textwidth}{!}
    {
        \begin{tabular}{c@{\hspace{.5\tabcolsep}}l|CCCCCCCCC} 
            \toprule 
            & & \mathbb{X} | \text{M} & \mathbb{X} | \text{S} & \mathbb{X} | \text{T} & \mathbb{X} | \text{MS} & \mathbb{X} | \text{MT} & \mathbb{X} | \text{ST} & \mathbb{X} | \text{MST}\\
            \midrule
            \multirow{2}{*}{MVAE} & vanilla & -8.61 \pm 1.18 & -10.26 \pm 1.14 & -8.17 \pm 1.30 & -8.18 \pm 1.26 & -1.51 \pm 0.11 & -7.41 \pm 1.38 & -1.02 \pm 0.11 \\
            & ours & -8.07 \pm 0.88 & -9.89 \pm 0.62 & \mathbf{-6.94 \pm 1.19} & -7.44 \pm 0.64 & -1.42 \pm 0.05 & -6.12 \pm 1.69 & -0.95 \pm 0.00 \\
            \midrule
            \multirow{2}{*}{MMVAE} & vanilla  & -2.30 \pm 0.21 & \mathbf{-2.18 \pm 0.09} & -1.19 \pm 0.00 & \mathbf{-2.24 \pm 0.11} & -1.75 \pm 0.10 & \mathbf{-1.68 \pm 0.04} & \mathbf{-1.89 \pm 0.07} \\
            & ours  & -2.59 \pm 0.40 & -2.31 \pm 0.13 & \mathbf{-1.19 \pm 0.00} & -2.45 \pm 0.23 & -1.89 \pm 0.20 & -1.75 \pm 0.06 & -2.03 \pm 0.15 \\
            \midrule
            \multirow{2}{*}{MoPoE} & vanilla & -1.93 \pm 0.01 &-2.06 \pm 0.04 &-1.19 \pm 0.00 &-1.76 \pm 0.00 &-1.18 \pm 0.00 &-1.04 \pm 0.00 &-1.03 \pm 0.05 \\
            & ours & -1.92 \pm 0.03 &-2.09 \pm 0.05 &\mathbf{-1.19 \pm 0.00} &-1.75 \pm 0.02 &-1.17 \pm 0.01 &\mathbf{-1.03 \pm 0.00} &-1.00 \pm 0.01\\
            \bottomrule
        \end{tabular}
    }
\end{table}

\begin{table}[h]
    \centering
     \caption{Log-likelihood of the joint generative model, conditioned on the variational posterior of subsets of the modalities. 
     Results report on IWAE as loss function and are averaged over 5 different seeds. Each log-likelihood is divided by the dimensionality of its modality before adding them up, to better reflect improvement across modalities.} \label{tab:log-lik-iwae}
    \resizebox{\textwidth}{!}
    {
        \begin{tabular}{c@{\hspace{.5\tabcolsep}}l|CCCCCCCCC} 
            \toprule 
            & & \mathbb{X} | \text{M} & \mathbb{X} | \text{S} & \mathbb{X} | \text{T} & \mathbb{X} | \text{MS} & \mathbb{X} | \text{MT} & \mathbb{X} | \text{ST} & \mathbb{X} | \text{MST}\\
            \midrule
            \multirow{2}{*}{MVAE} & vanilla & -8.62 \pm 0.40 & -10.68 \pm 0.98 & -6.78 \pm 1.66 & -8.28 \pm 0.47 & -1.83 \pm 0.14 & -6.08 \pm 1.36 & -1.15 \pm 0.03 \\
            & ours & \mathbf{-8.26 \pm 0.29} & \mathbf{-9.51 \pm 0.25} & \mathbf{-2.83 \pm 0.54} & \mathbf{-7.74 \pm 0.15} & \mathbf{-1.35 \pm 0.03} & \mathbf{-1.96 \pm 0.33} & \mathbf{-0.94 \pm 0.00} \\
            \midrule
            \multirow{2}{*}{MMVAE} & vanilla & \mathbf{-1.98 \pm 0.06} & -2.74 \pm 0.32 & -1.27 \pm 0.00 & -2.36 \pm 0.15 & \mathbf{-1.62 \pm 0.03} & -2.00 \pm 0.16 & -1.99 \pm 0.10 \\
            & ours  & -2.44 \pm 0.03 & \mathbf{-2.45 \pm 0.12} & \mathbf{-1.26 \pm 0.00} & -2.44 \pm 0.06 & -1.85 \pm 0.01 & \mathbf{-1.86 \pm 0.06} & -2.05 \pm 0.04 \\
            \midrule
            \multirow{2}{*}{MoPoE} & vanilla & \mathbf{-1.97 \pm 0.00} & -2.55 \pm 0.12 & -1.27 \pm 0.00 & \mathbf{-5.85 \pm 0.52} & -1.25 \pm 0.00 & -1.05 \pm 0.00 & -1.02 \pm 0.00 \\
            & ours & -2.01 \pm 0.01 & -2.61 \pm 0.11 & \mathbf{-1.27 \pm 0.00} & -7.07 \pm 0.06 & \mathbf{-1.24 \pm 0.01} & -1.05 \pm 0.00 & \mathbf{-1.01 \pm 0.00} \\
            \bottomrule
        \end{tabular}
    }
\end{table}

\begin{table}[h]
    \centering
     \caption{Log-likelihood of the joint generative model, conditioned on the variational posterior of subsets of the modalities. 
     Results report on SIWAE as loss function and are averaged over 5 different seeds. Each log-likelihood is divided by the dimensionality of its modality before adding them up, to better reflect improvement across modalities.} \label{tab:log-lik-siwae}
    \resizebox{\textwidth}{!}
    {
        \begin{tabular}{c@{\hspace{.5\tabcolsep}}l|CCCCCCCCC} 
            \toprule 
            & & \mathbb{X} | \text{M} & \mathbb{X} | \text{S} & \mathbb{X} | \text{T} & \mathbb{X} | \text{MS} & \mathbb{X} | \text{MT} & \mathbb{X} | \text{ST} & \mathbb{X} | \text{MST}\\
            \midrule
            \multirow{2}{*}{MVAE} & vanilla & -8.82 \pm 0.63 & -10.13 \pm 0.29 & -7.17 \pm 1.73 & -8.52 \pm 0.74 & -2.23 \pm 1.31 & -5.94 \pm 1.35 & -1.15 \pm 0.02 \\
            & ours  & -8.41 \pm 0.14 & -10.52 \pm 0.53 & -5.65 \pm 1.76 & -8.13 \pm 0.17 & -1.46 \pm 0.07 & -5.49 \pm 1.87 & \mathbf{-1.01 \pm 0.01} \\
            \midrule
            \multirow{2}{*}{MMVAE} & vanilla & \mathbf{-2.01 \pm 0.08} & -4.04 \pm 0.43 & -3.25 \pm 0.94 & -3.02 \pm 0.20 & -2.63 \pm 0.49 & -3.64 \pm 0.25 & -3.10 \pm 0.18 \\
            & ours & -2.34 \pm 0.36 & -3.22 \pm 1.28 & -3.03 \pm 1.07 & -2.78 \pm 0.59 & -2.69 \pm 0.54 & \mathbf{-3.12 \pm 0.43} & \mathbf{-2.86 \pm 0.22} \\
            \midrule
            \multirow{2}{*}{MoPoE} & vanilla & -6.29 \pm 2.75 & -10.02 \pm 0.43 & -3.80 \pm 1.48 & -7.51 \pm 0.93 & -1.59 \pm 0.35 & -4.97 \pm 2.68 & -3.15 \pm 3.78 \\
            & ours & -8.01 \pm 0.42 & -10.16 \pm 0.55 & -2.96 \pm 1.11 & -7.26 \pm 0.28 & -1.34 \pm 0.01 & \mathbf{-2.30 \pm 0.91} & -0.99 \pm 0.02 \\
            \bottomrule
        \end{tabular}
    }
\end{table}
}

\end{document}